%% file: main.tex
\documentclass[sigconf, nonacm]{acmart}

\usepackage{hyperref}
\hypersetup{
    colorlinks,
    linkcolor={blue!50!black},
    citecolor={blue!50!black},
    urlcolor={blue!80!black}
}

\usepackage{url}
\usepackage{amsmath}

\usepackage[table]{xcolor}

\usepackage[T1]{fontenc}
\usepackage{graphicx}
\usepackage{booktabs}
\usepackage{listings}
\usepackage{multirow}

\usepackage{amssymb}

\usepackage[most]{tcolorbox}
\usepackage{makecell}

\colorlet{lightblue}{blue!15}
\colorlet{lightorange}{orange!13}

\definecolor{codegreen}{rgb}{0,0.6,0}
\definecolor{codegray}{rgb}{0.5,0.5,0.5}
\definecolor{backcolour}{RGB}{255,255,255}
\definecolor{emph}{RGB}{166,88,53}
\definecolor{nightblue}{RGB}{9,49,105}
\definecolor{keywords}{RGB}{207,33,46}
\definecolor{lightpurple}{RGB}{130,81,223}
\lstdefinelanguage{json}{
    basicstyle=\fontsize{7}{8}\ttfamily,
    stepnumber=1,
    numbersep=8pt,
    showstringspaces=false,
    breaklines=true,
    frame=lines,
    backgroundcolor=\color{backcolour},   
    literate=
     *{0}{{{\color{numb}0}}}{1}
      {1}{{{\color{numb}1}}}{1}
      {2}{{{\color{numb}2}}}{1}
      {3}{{{\color{numb}3}}}{1}
      {4}{{{\color{numb}4}}}{1}
      {5}{{{\color{numb}5}}}{1}
      {6}{{{\color{numb}6}}}{1}
      {7}{{{\color{numb}7}}}{1}
      {8}{{{\color{numb}8}}}{1}
      {9}{{{\color{numb}9}}}{1}
      {:}{{{\color{punct}{:}}}}{1}
      {,}{{{\color{punct}{,}}}}{1}
      {\{}{{{\color{delim}{\{}}}}{1}
      {\}}{{{\color{delim}{\}}}}}{1}
      {[}{{{\color{delim}{[}}}}{1}
      {]}{{{\color{delim}{]}}}}{1},
}

\lstdefinestyle{mystyle}{
    escapeinside={(*}{*)},
    backgroundcolor=\color{backcolour},   
    commentstyle=\color{codegreen},
    keywordstyle=\color{keywords},
    stringstyle=\color{nightblue},
    basicstyle=\fontsize{8}{9}\ttfamily,
    breakatwhitespace=true,         
    breaklines=true,                 
    captionpos=b,                    
    keepspaces=true,                 
    numberstyle=\tiny\color{codegray},
    numbersep=2pt,                  
    showspaces=false,                
    showstringspaces=false,
    showtabs=false,                  
    tabsize=2,
    emph={dspy},
    emphstyle={\color{lightpurple}},
    linewidth=1\columnwidth,
    frame=tb,   
    xrightmargin=0pt,
    xleftmargin=0.23cm,
    aboveskip=0.2cm,
    belowskip=0.1cm,
    otherkeywords={OFFSET},
}

\usepackage[table]{xcolor}
\definecolor{lightgray}{gray}{0.95}

\usepackage[bottom]{footmisc} % To fix paragraph

\newcommand{\inlinesql}[1]{\lstinline[language=SQL,showstringspaces=false]{#1}}

\newcommand{\inlinesqlblack}[1]{{\ttfamily\small #1}}

\lstnewenvironment{sqlcode}
{\lstset{language=SQL,breaklines=true,showstringspaces=false}}
{}

\lstnewenvironment{pythoncode}
{\lstset{language=python,breaklines=true,showstringspaces=false}}
{}

\usepackage{fancyvrb}
\usepackage{xcolor}
 
\newcommand{\blue}[1]{\textcolor{blue}{#1}} 

\usepackage{algorithmicx}
\usepackage{algorithm}
\usepackage{algpseudocode}

\makeatletter
\renewcommand\subsubsection{\@startsection{subsubsection}{3}{\z@}%
  {-10\p@ \@plus -4\p@ \@minus -2\p@}%
  {4\p@}%
  {\normalfont\normalsize\bfseries}} % Changed from \itshape

\renewcommand\paragraph{\@startsection{paragraph}{4}{\z@}%
  {\z@}%
  {-1em}%
  {\normalfont\normalsize\bfseries}} % Changed from \itshape
\makeatother

\definecolor{goodgreen}{RGB}{198, 239, 206}
\definecolor{badred}{RGB}{255, 199, 206}
\usepackage{fontawesome5}

\usepackage{siunitx}
\definecolor{bestgreen}{RGB}{198,239,206}
\definecolor{tiecolor}{RGB}{255,255,191}
\definecolor{rowgray}{RGB}{245,245,245}
\usepackage{tikz}

\usepackage{graphicx}
\newcommand{\imageemoji}{\raisebox{-0.2em}{\includegraphics[height=1.2em]{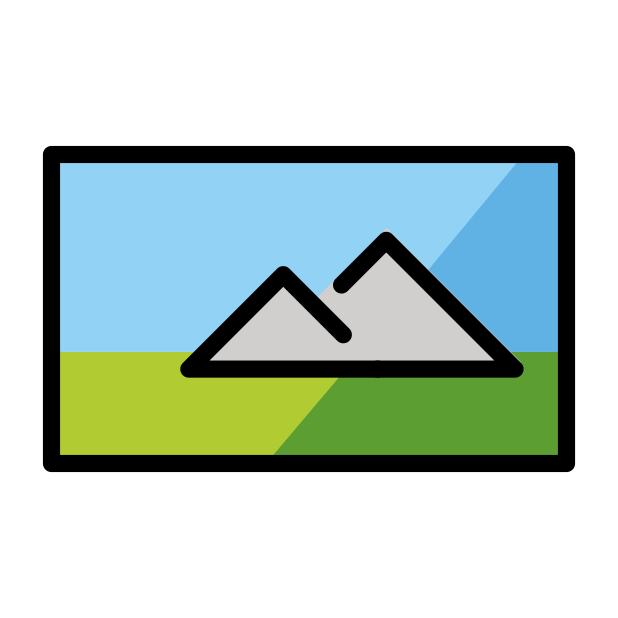}}}
\newcommand{\textemoji}{\raisebox{-0.2em}{\includegraphics[height=1.2em]{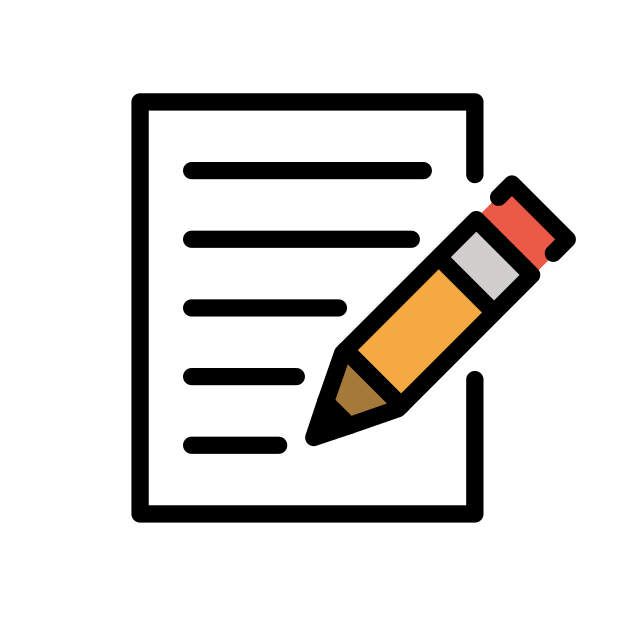}}}
\newcommand{\audioemoji}{\raisebox{-0.2em}{\includegraphics[height=1.2em]{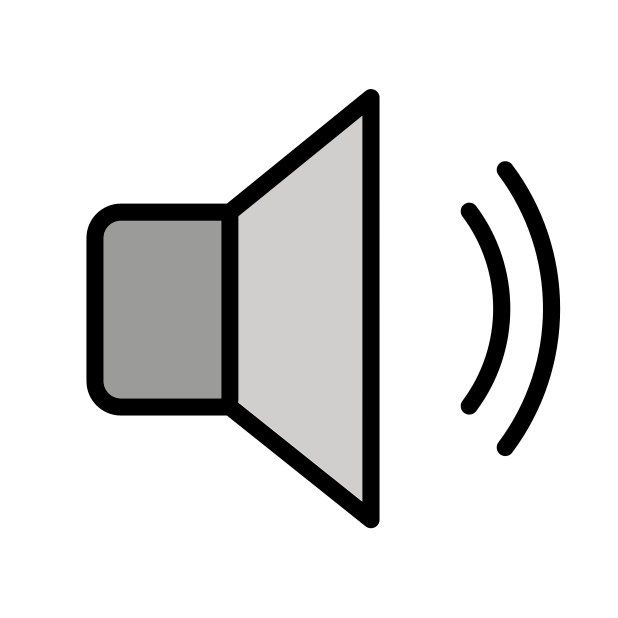}}}
\usepackage{caption}
\usepackage{subcaption}
\usepackage{placeins}

\begin{document}
\title{Large Databases Need Small, Open-Weight Language Models}

%%
%% The "author" command and its associated commands are used to define the authors and their affiliations.
\author{Parker Glenn}
\affiliation{%
  \institution{Capital One}
}
\email{parker.glenn@capitalone.com}

\author{Alfy Samuel}

\affiliation{%
  \institution{Capital One}
}
\email{alfy.samuel@capitalone.com}

\begin{abstract}

Language model systems built around proprietary APIs often operate on a token-based cost model. This becomes prohibitively expensive in the context of large databases, where LM-enhanced relational operators can incur costs exceeding \$10,000 for a single set of experiments, hindering thorough research and practical deployment.
In this paper, we demonstrate that quantized, open-weight models running locally on just 16GB of VRAM can match or exceed the accuracy of closed-source counterparts at lower latency and a fraction of the price, challenging the prevailing assumption that closed-source LM APIs are necessary for effective LM-database integration. 
We present and analyze the key system optimizations required to efficiently deploy these open-weight models within an LM-DB system.
By integrating these local models into the \textsc{BlendSQL} v0.1.0 framework, we demonstrate a 390x reduction in overall costs and 3.8x reduction in latency compared to a proprietary LM API. We make our code available at \href{https://github.com/CapitalOne-Research/play-by-the-type-rules/tree/main/sembench}{\textcolor{blue!50}{https://github.com/CapitalOne-Research/play-by-the-type-rules/tree/main/sembench}}.

\end{abstract}

\maketitle

\input{sections/1_introduction.tex}

\input{sections/2_system_overview}

\input{sections/3_experiments}

\input{sections/4_results}

\input{sections/5_related_work}

\input{sections/6_conclusion}

% \begin{acks}
%  This work was supported by the [...] Research Fund of [...] (Number [...]). Additional funding was provided by [...] and [...]. We also thank [...] for contributing [...].
% \end{acks}

%\clearpage

\bibliographystyle{ACM-Reference-Format}
\bibliography{sample}

\input{sections/7_appendix}

\end{document}

%% file: sections/1_introduction.tex
\section{Introduction}

Relational databases remain the ubiquitous choice for storing structured information. 
Combined with the increasing popularity of language models, a research question has emerged: what is the optimal method for combining the flexible reasoning capabilities of language models with the deterministic and reliable processing of traditional structured query languages? 
This notion of an \textit{optimal} approach can be decomposed into three key dimensions: cost, latency, and quality. 
To illustrate, consider an auto repair shop with a database of customer complaints containing columns \inlinesqlblack{summary} and \inlinesqlblack{date}. A user might seek all complaints describing an engine-related issue submitted before 2024. Determining whether a complaint is engine-related requires interpreting free-text summaries, which is out-of-scope for native SQL operators. A recent trend is to accomplish this via a call to a language model (LM): \inlinesql{SELECT summary FROM complaints WHERE LM('Is this engine-related?', summary) = TRUE AND date < '2024-01-01'}.
Queries containing LM functions may be made more complex by referencing multiple tables via \inlinesql{JOIN} clauses, injecting additional native-SQL conditions, or adding additional LM functions within a single expression context, all of which may change the terms of what makes a query plan \textit{optimal}. 
Importantly, the introduction of non-deterministic LM functions inherently changes the optimization landscape for database systems. 

A growing body of research shows that the quality of the software infrastructure wrapping language model calls can be as predictive of downstream task performance as the choice of base LM. In agentic coding flows, harness-level changes such as self-verification loops, context management, and loop detection, have produced large benchmark improvements without any change to the underlying model  \cite{lee2026meta,li2025deepagent,wang2024openhands}. In tool calling, structured generation systems have been shown to elevate open-source models to performance meeting or exceeding closed-source counterparts \cite{wang2025slot, geng2025jsonschemabench}. 
% In general reasoning tasks, optimal allocation of test-time compute can elevate performance of a small language model to that of a 14x larger model \cite{snell2024scaling}. 
These gains are driven not by innovations in pre/post-training or parameter scaling, but by embedding task-specific inductive bias into the systems that orchestrate inference. 

\begin{figure}[t!]
    \centering
     \includegraphics[scale=0.6]{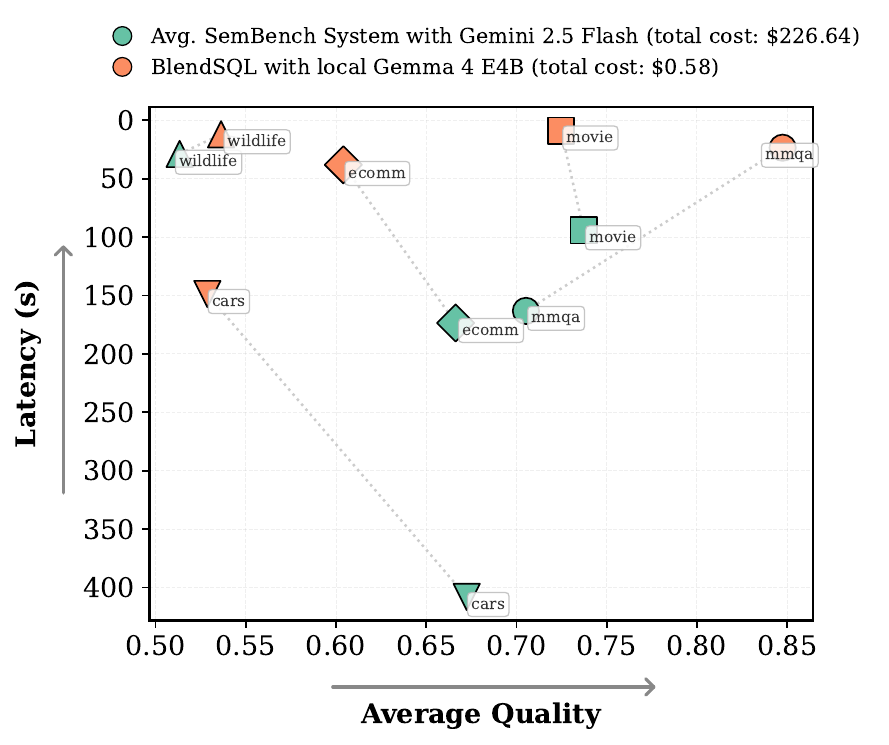}
    \caption{Plotting the quality, latency, and cost over the five scenarios in SemBench. Latency and quality are averaged across five total runs, and cost is computed as the cumulative total of all runs.}
    \label{fig:overall_summary}
\end{figure}

We argue that, in spite of recent progress made in hybrid LM-DB systems, there are still large cost, quality, and latency gains to be made with small, open-weight models by designing effective software infrastructure for hybrid LM-DB systems. 
To demonstrate this, we sample a recent body of work applying language models to relational databases via a UDF-style pattern. 
Of 21 total evaluated language models across 8 published works, only 6 are open-weight models \cite{su2026large,lin2026sema,shahbazi2026omnitqa,chowdhury2026diagnosis,khoja2025weaver,lao2025sembench,mang2026horrila,zhao2024hybrid}. 
Of the 6 evaluated open-weight models across published works, none are below 20B parameters. 
% In the absence of extreme quantization, these cannot be run on a single consumer-grade GPU and are prone to incurring large \$ costs over large databases.
While the cost of these proprietary, API-based models is often undisclosed, \citet{lao2025sembench} report total costs of \$10,167.80 in their published experiments using Gemini 2.5 Flash. 
Importantly, this cost is a subset of total experimental cost, as it does not include preliminary experiments leading up to final results. 
We show that not only are these costly proprietary models slow and unnecessary, but, within certain domains, they are strictly worse than a properly harnessed local open-weight model. 

Our contributions are the following: 

% \begin{itemize}
% \item We present \textsc{BlendSQL} v0.1.0, a highly expressive LM-DB system whose core polymorphic functions are capable of natively handling all 54 complex queries across the multimodal SemBench benchmark.
% \item We demonstrate that \textsc{BlendSQL} circumvents the linear memory scaling deficiencies of existing baseline systems. By cleanly delegating intermediate relational operations to the native DBMS, it maintains a near-constant memory footprint, enabling small, quantized open-weight models to run efficiently on a single 16GB GPU.
% \item We show that our system matches the quality of closed-source API alternatives at a 342× reduction in overall cost and a 4.8× reduction in latency. We ablate these gains to reveal that constrained decoding yields up to +0.85 quality improvement with negligible latency overhead, while early exiting reduces latency by up to 65\% on \inlinesql{LIMIT}-heavy workloads.
% \item We identify a modality gap in current small open-weight models: while competitive with closed counterparts on text and image inputs, they fall significantly behind on audio, suggesting that audio understanding remains a bottleneck for cost-effective local deployment.
% \end{itemize}

\begin{itemize}
\item We present \textsc{BlendSQL} v0.1.0, a LM-DB system that combines query-level optimizations with constrained decoding to enable small, quantized open-weight models running on a single 16GB GPU to achieve a win-or-tie rate of 57\% against closed-source alternatives at 390× lower cost and 3.8× lower latency on the SemBench benchmark.
\item We demonstrate that \textsc{BlendSQL} circumvents the linear memory scaling deficiencies of existing baseline systems at large database scales, maintaining a near-constant memory footprint.
% \item We ablate each system optimization in isolation, showing that constrained decoding yields up to +0.85 quality improvement with negligible latency overhead, and that early exiting reduces latency by up to 65\% on \inlinesql{LIMIT}-heavy workloads.
\item We identify a modality gap in current small open-weight models: while competitive with closed counterparts on text, they fall behind on image and audio, suggesting that multi-modal understanding remains a bottleneck for cost-effective local deployment.
\item We argue for a future of LM-DB research built on open-weight models, ensuring accessibility and full reproducibility of experimental results. 
\end{itemize}

% Open-weight models are preferable to closed-weight models for the following reasons:

% \begin{itemize}
%     \item Closed-weight APIs are constantly changing, making reproducibility impossible.
%     \item Closed-weight APIs enforce arbitrary rate limiting, inhibiting the ability to massively parallelize inference calls.
%     \item Closed-weight APIs typically operate on a token-based cost model. In the presence of large databases, this causes the cost of frequent iterations and experiments to skyrocket.
% \end{itemize}

%% file: sections/2_system_overview.tex
\section{System Overview} \label{sec:system_overview}

\subsection{Terminology}

Several terms have recently been introduced to describe systems that combine traditional relational operators with AI-powered operators. \citet{lao2025sembench} introduce \textit{semantic query processing engines (SQPEs)} to describe systems that extend relational algebra with natural language \textit{semantic operators}. Similarly, \citet{su2026large} use \textit{LLM-Enhanced Relational Operators (LROs)} to denote components that enhance relational processing by invoking a language model. More broadly, \citet{yan2025dbms} categorize this intersection of language models and databases as \textit{DBMS–LLM systems}, a space further refined by \citet{zhou2023serving}, who formally distinguish the \textit{UDF (user-defined function)-centric} integration pattern presently relevant to our work.

We adopt the umbrella term \textit{LM-DB systems} to refer to systems that integrate language model functions with existing structured data manipulation interfaces such as Pandas and SQL. We use the broader term ``language model'' (LM) to avoid restricting our scope to models of any particular scale. Ultimately, each system we study is a UDF-centric system, and can be decomposed into two components: (1) a \textit{query optimizer}, which routes subsets of data to LM functions, and (2) the \textit{LM function implementation} itself, encompassing the prompts, generation parameters, and post-processing logic used at inference time. We describe each component of our proposed system below. 

\subsection{\textsc{BlendSQL} v0.1.0 Implementation} \label{sec:blendsql_overview}

\textsc{BlendSQL} is a query language that compiles to SQL \citep{glennblendsql}. It allows for combining traditional SQL operators with generalizable language model (LM) functions capable of unstructured reasoning. Each \textsc{BlendSQL} LM function is denoted by double curly braces, ``\texttt{\{\{}'' and ``\texttt{\}\}}''. Using a pre-determined prompt template, a response is generated from a local or remote language model with optional type-constraints to yield a function output. This function output is then integrated into the wider SQL query using the \texttt{sqlglot} parser \cite{sqlglot}. Once all LM functions have run, a final compiled SQL query is executed against the native DBMS to yield the final result. Currently, SQLite, DuckDB, and PostgreSQL backends are supported. 
% Given this function output, an abstract-syntax tree (AST) transformation rule is applied to the original query AST to yield a syntactically valid SQL query, which can be executed by the native database execution engine. Certain functions, such as the row-wise \textsc{llmmap} function, rely on the creation of temporary tables to integrate function outputs into the wider SQL query. This level of integration with the DBMS allows for the scaling of \textsc{BlendSQL} to any database which supports the creation of temporary tables that expire upon session disconnect. Currently, SQLite, DuckDB, and PostgreSQL backends are supported. 
In this work, we introduce \textsc{BlendSQL} v0.1.0. Since the original v0.0.0 release in \citet{glennblendsql}, the following changes have been made:

\begin{itemize} 
  \item An improved type inference system that constrains LM generations to valid datatypes.
  \item Support for image and audio datatypes as arguments to LM functions.
  \item Query optimizations including cascade filtering and early exiting.
  \item Faster aggregation of in-memory LM function outputs via \texttt{polars} \cite{polars}.
  % \item Batched inputs and outputs for \textsc{llmmap} functions replaced for the more commonly used single-input-single-output pattern.
\end{itemize}

\subsection{LM Functions}\label{sec:lm_functions}

Unlike other LM-DB systems that rely on numerous operators coupled with specific prompt templates (e.g., \texttt{sem\_map}, \texttt{sem\_filter}, \texttt{sem\_extract}, \texttt{sem\_topk}), \textsc{BlendSQL} implements a small number of general-purpose polymorphic functions defined by their input/output cardinality. In this work, we focus on two core polymorphic functions and demonstrate how they enable complex reasoning patterns such as ranking, RAG, and entity linking. For comprehensive descriptions and function signatures for all available functions, refer to the actively maintained online documentation\footnote{\url{https://parkervg.github.io/blendsql/reference/functions/}}.

\paragraph{Constrained Decoding} All \textsc{BlendSQL} v0.1.0 LM functions accept a \texttt{return\_type} argument, which is either specified explicitly or inferred implicitly from query syntax \cite{glenn2025play}. This \texttt{return\_type} is then used to perform constrained decoding, guaranteeing adherence to datatype structures defined via a context-free grammar. When provided, the \texttt{options} argument behaves similarly - constrained decoding will be used to restrict LM generations to a value in the option set. If the \texttt{return\_type} is a collection, the \texttt{quantifier} argument can be used to apply a regular expression-style modifier to restrict the number of generated items (e.g. \texttt{\{3\}} for ``exactly three'',  \texttt{\{1,5\}} for ``between one and five'').

\paragraph{One-Shot Prompting}

All \textsc{BlendSQL} v0.1.0 LM functions employ type-aligned one-shot prompting. From a pre-defined pool, the system selects a single example that matches the current function's expected return type.

\subsubsection{\textsc{llmqa}}\label{sec:llmqa}
The \textsc{llmqa} function is an aggregate function which transforms a subset of data into a single-cell output. This output can be a single scalar type (\texttt{return\_type='str'}), or a collection (\path{return_type='List[str]'}). 

\paragraph{Example 1: \textsc{llmqa} for top-k ranking}
\mbox{} 
\begin{lstlisting}[language=SQL,breaklines=true,showstringspaces=false]
SELECT * FROM (
    VALUES {{
        LLMQA(
            'Select the 3 most 
            negative reviews.',
            options=(
                SELECT reviewText 
                FROM Reviews 
                WHERE reviewYear = 2025
            ),
            return_type='List[str]',
            quantifier='{3}'
        )
    }} 
) AS rankedReviews(review1, review2, review3)
\end{lstlisting}

\paragraph{Example 2: \textsc{llmqa} for unstructured-structured linking}
\mbox{} 
\begin{lstlisting}[language=SQL,breaklines=true,showstringspaces=false]
SELECT name FROM state_flowers
WHERE state = {{
    LLMQA(
        'Which state is known as 
        ''The Golden State''?',
        context=(SELECT * FROM documents)
    )
}}
\end{lstlisting}

% \begin{pythoncode}
% def LLMQA(
%     question: str,
%     *context: Optional[Union[ColumnRef, Subquery] # Can be a BlendSQL subquery, or a `{tablename}.{colname}` reference
% ):
%     ...
% \end{pythoncode}

\subsubsection{\textsc{llmmap}}\label{sec:llmmap}

The \textsc{llmmap} function is a row-wise function that takes a natural language question and one or more column names. For each row, the corresponding values $v_1, \dots, v_k$ are inserted into the prompt alongside the question $Q$ and passed to the underlying LM, which returns the result of $f(Q, v_1, \dots, v_k)$. As with \textsc{llmqa}, $f$ returns either a single scalar type (\mbox{\texttt{return\_type=`str'}}) or a collection (\mbox{\texttt{return\_type=`List[str]'}}). 

\paragraph{Refined \textsc{llmmap} Prompting Pattern}
Previously in \textsc{BlendSQL} <v0.1.0, the \textsc{llmmap} function would run with a default \texttt{batch\_size} argument of 5, which determined the number of input values to pass in a single prompt. The language model was then expected to generate the mapped outputs for each value, separated by a semicolon. For example, given the input:
\begin{verbatim}
What is the capital of the country?
France
Croatia
Canada
Australia
Ukraine
\end{verbatim}
the LM would be expected to generate the string \texttt{Paris;\allowbreak Zagreb;\allowbreak Ottawa;\allowbreak Canberra;\allowbreak Kyiv}. While this approach greatly decreased token usage, it also led to sub-optimal accuracy in many studies \cite{lin2026sema,su2026large}. We adopt the more traditional single-in, single-out prompting format (i.e.\ \texttt{batch\_size=1}) as the default for \textsc{BlendSQL} v0.1.0.

\paragraph{Example 3: \textsc{llmmap} for filtering}
\mbox{} 
\begin{lstlisting}[language=SQL,breaklines=true,showstringspaces=false]
SELECT * FROM Reviews 
WHERE {{
LLMMap(
    'Does this review have a positive sentiment?', 
    reviewText
)
}} = TRUE
\end{lstlisting}

\paragraph{Example 4: \textsc{llmmap} for classification and grouping}
\mbox{} 
\begin{lstlisting}[language=SQL,breaklines=true,showstringspaces=false]
SELECT GROUP_CONCAT(Name, ', ') AS 'Names',
{{
    LLMMap(
        'In which century was
        this person born?',
        p.Name,
        options=('1800s', '1900s', '2000s')
    )
}} AS Born
FROM People
GROUP BY Born
\end{lstlisting}

% \begin{pythoncode}
% def LLMMap(
%     question: str,
%     values: ColumnRef # {tablename}.{colname}
% ):
%     ...
% \end{pythoncode}

% \paragraph{Performing Joins}

% \textsc{BlendSQL} implements a semantic left-\inlinesql{JOIN} implementation via \textsc{llmjoin}. 

\subsection{Query Optimization} \label{sec:query_optimization}
Optimizing user-defined functions (UDFs) is a notoriously difficult task. A large body of research has explored this in the general setting \cite{arch2024key,foufoulas2023efficient,wehrstein2025graceful,foufoulas2025udfbench}: given some user-provided function of unknown computational cost, what is the optimal method of execution within a wider SQL query? Notably, when embedding language models into SQL, the problem space is different in an important way: Given prior knowledge that a function $f_{LM}$ calls an LM while other functions $f$ do not, the cost calculation $C(f(D))\ll C(f_{LM}(D)) $ typically holds for any data subset $D$. Then, given \textsc{llmqa} calls translate to exactly one LM generation call and \textsc{llmmap} potentially $ > 1$, we derive the following cost model by abstracting away $D$: 
\begin{equation} \label{eq:cost_model}
C(f) \ll C(\text{\textsc{llmqa}}) < C(\text{\textsc{llmmap}})
\end{equation}
We implement a rule-based optimizer built on this heuristic cost model. When executing a program, the query is parsed into an abstract syntax tree (AST) using \texttt{sqlglot} \cite{sqlglot}. Each \inlinesql{SELECT} expression is iteratively processed via depth-first search, and the following strategy is applied:

\begin{enumerate}
  \item \textbf{Pre-filtering}: Blocking operators are removed (\inlinesql{ORDER BY}, \inlinesql{GROUP BY}, etc.) and LM function nodes are replaced with the constant \inlinesql{TRUE}, allowing the underlying DBMS to execute all vanilla SQL predicates first. The \inlinesql{SELECT} argument is modified to only select those columns used by LM functions, minimizing I/O overhead.
  \item \textbf{Materialization}: The filtered query is executed against the DBMS and written to a temporary table, effectively pushing down all non-LM operations.
  \item \textbf{LM Function Execution}: Following the standard SQL order of operations 
(\inlinesql{FROM/JOIN}$\rightarrow$\inlinesql{WHERE}$\rightarrow$\inlinesql{GROUP BY}, etc.), the LM functions are executed, reading from the filtered temporary tables to fetch inputs.
\end{enumerate}

The \inlinesql{TRUE} substitution in Step 1 ensures the optimizer honors disjunctive conditions between LM functions and vanilla SQL functions (\inlinesql{WHERE} $f_{LM}()$ \inlinesql{OR f()} becomes \inlinesql{WHERE TRUE OR f()}, preventing an overly-eager pre-filter via \inlinesql{f()}).

\paragraph{Cascade Filtering}
\textsc{BlendSQL} implements cascade filtering for conjunctive predicates involving LM functions. The LM predicates are sequenced into a pipeline of operators, and subsequent functions only process rows of data which satisfied all preceding predicates. This avoids redundant, expensive calls for rows already disqualified by a prior LM-based filter. This logic applies when two or more LM functions appear within a \inlinesql{WHERE} clause, provided that no disjunctions (\inlinesql{OR} relationships) exist between them. Currently, our optimizer prioritizes \textsc{llmqa} before \textsc{llmmap} based on the cost model in Eq. \ref{eq:cost_model}. Future work may refine the ordering of multiple \textsc{llmmap} calls by estimating and prioritizing the most selective predicates first.

\paragraph{Early Exiting}
In expressions where a \inlinesql{LIMIT} keyword is used, \textsc{llmmap} execution may terminate as soon as the required number of rows is satisfied. This optimization is only applied when the expression contains no blocking operators (e.g. \inlinesql{ORDER BY}, \inlinesql{GROUP BY}, etc.) that would require a full scan of the LM function's output. Building on the \texttt{AsyncOpenAI} API,  we utilize streaming to ensure no unnecessary tokens are generated: as soon as the exit condition is met, a cancel event is issued and all outstanding asynchronous generation requests are terminated.

\paragraph{Early Deduplication of Database Values}

Before passing a sequence of values to an \textsc{llmmap} function, the input set is deduplicated via a \inlinesql{SELECT DISTINCT} clause, and the mapped outputs are then aligned to the original table via a \inlinesql{LEFT JOIN} operation. Consider the query \inlinesql{SELECT name FROM People WHERE \{\{LLMMap('Are they an NBA player?', name)\}\} = TRUE} over a table with $N = 10^6$ rows but only $|\text{dom}(name)| \ll 10^6$ distinct name values. Without deduplication, each row would result in a separate LM invocation; with early deduplication, the number of generation calls is reduced to $|\text{dom}(name)|$, and results are broadcast back to all matching rows via the join. By paying a relatively small upfront cost of deduplicating and re-joining to the source table, this optimization yields a reduction in LM calls proportional to the duplication factor $\rho = N / |\text{dom}(name)|$. Given the aforementioned cost heuristic $C(f(D))\ll C(f_{LM}(D)) $, this translates to substantially lower latency and cost. This optimization applies for both single-column and multi-column inputs to the \textsc{llmmap} function.

%% file: sections/3_experiments.tex
\section{Experiments}

\subsection{SemBench}

SemBench is a benchmark designed to evaluate the performance of hybrid LM-DB systems. The benchmark can be viewed as an extension of TPC-H \cite{tpc_h_standard} and TPC-DS \cite{tpc_ds_standard}, with a specific emphasis on the cost, quality, and latency of queries containing LM functions. It consists of five scenarios, requiring LM functions to operate over a mix of text, image, and audio data. In the example below from the \textsc{ecomm} scenario, a system must orchestrate calls to a multi-modal language model to determine the color of products, given local image files.

\mbox{} 
\begin{lstlisting}[language=SQL,breaklines=true,showstringspaces=false]
SELECT
i.id as id,
{{
  LLMMap(
    'What is the primary color of the product in this image?',
    i.local_image_path
  )
}} AS category
FROM styles_details s
JOIN image_mapping i on s.id = i.id
WHERE s.baseColour IN 
('Black', 'Blue', 'Red', 'White', 
'Orange', 'Green')
\end{lstlisting}

\begin{figure*}[h]
    \centering
     \includegraphics[scale=0.4]{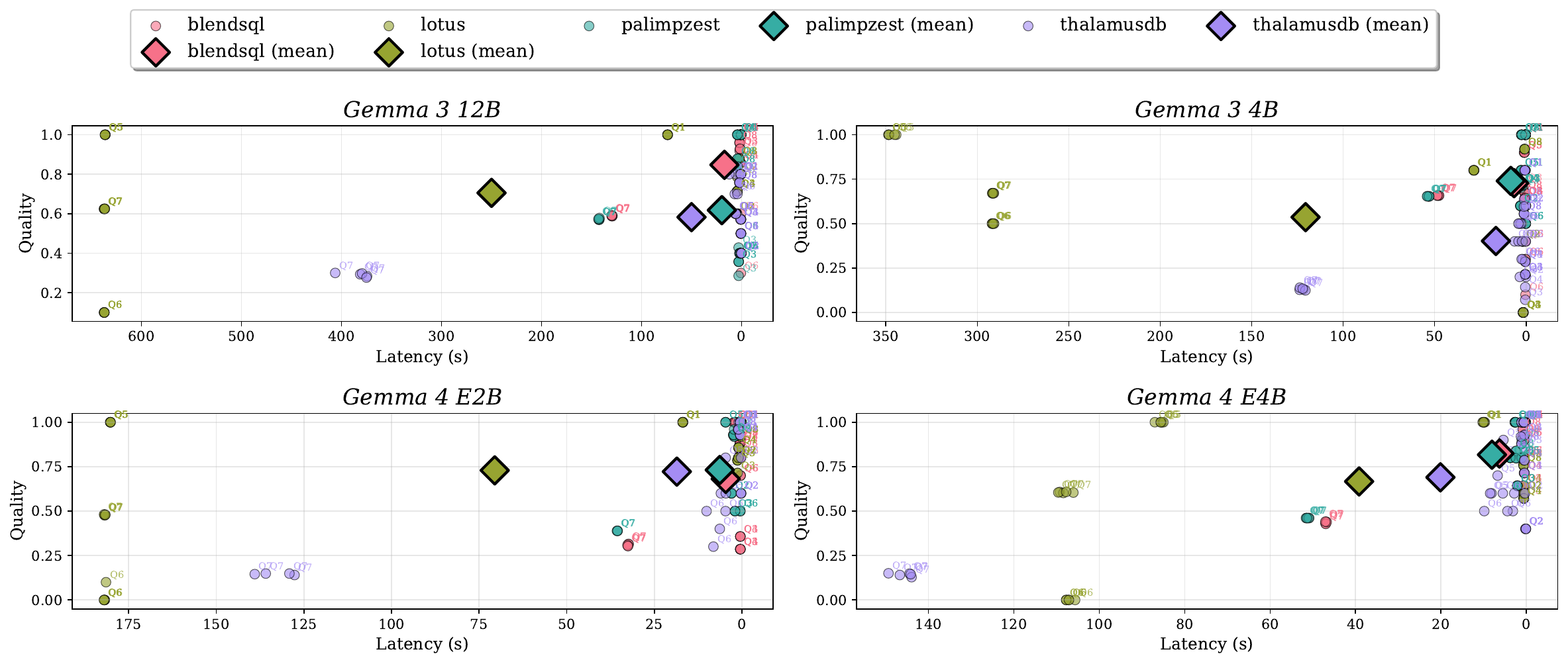}
    \caption{Benchmarking the quality and latency of various LM-DB systems with open-weight LMs on the text-only \textsc{movie} scenario from SemBench. All systems are run with a max concurrency of 32.}
    \label{fig:comparison_against_other_systems}
\end{figure*}

\subsection{Setup}\label{sec:model_inference}

\paragraph{Model Inference}
All experiments with open-weight models are run locally with an NVIDIA RTX 5080 GPU with 16GB VRAM, 64GB of RAM, and an AMD Ryzen 5 7600X processor. We experiment with the Gemma 3 \cite{gemmateam2025gemma3technicalreport} and Gemma 4 \cite{gemma4} families of models for their strong performance at sizes that fit within 16GB VRAM. For brevity, we use the shorthands in Table~\ref{tab:models} to refer to the quantized HuggingFace checkpoints used in our experiments. ~
\begin{table}[h]
  \centering
  \caption{Model shorthands and HuggingFace identifiers.}
  \label{tab:models}
  \begin{tabular}{ll}
    \toprule
    \textbf{Shorthand} & \textbf{HuggingFace Model} \\
    \midrule
    Gemma 4 E2B & \href{https://huggingface.co/prithivMLmods/gemma-4-E2B-it-FP8}{prithivMLmods/gemma-4-E2B-it-FP8} \\
    Gemma 4 E4B & \href{https://huggingface.co/prithivMLmods/gemma-4-E4B-it-FP8}{prithivMLmods/gemma-4-E4B-it-FP8} \\
    Gemma 3 4B  & \href{https://huggingface.co/RedHatAI/gemma-3-4b-it-quantized.w4a16}{RedHatAI/gemma-3-4b-it-quantized.w4a16} \\
    Gemma 3 12B & \href{https://huggingface.co/RedHatAI/gemma-3-12b-it-quantized.w4a16}{RedHatAI/gemma-3-12b-it-quantized.w4a16} \\
    \bottomrule
  \end{tabular}
\end{table}
~We use \texttt{vLLM==0.21.0} for model inference \cite{kwon2023efficient}. The \texttt{llguidance} engine is used for all constrained decoding capabilities \cite{guidance}. All experiments are run five times and their averages reported.

\paragraph{System Versions}

We use \texttt{blendsql==0.1.26}, \texttt{lotus-ai==1.1.4}, \texttt{palimpzest==0.8.2}, and \texttt{thalamusdb==0.1.15} in our experiments.

\paragraph{Cost Calculations}

While on-demand GPU pricing is highly volatile, hourly rental costs for an NVIDIA RTX 5080 range from \$0.14 to \$0.18 as of April 2026 \cite{vastaigpupricing, saladgpupricing}. We use the upper-end price of \$0.18 per hour for cost reporting of local models. For Gemini 2.5 Flash, we follow the cost reporting of \citet{lao2025sembench}: \$0.30 per million input text, image, and video tokens, \$1.00 per million input audio tokens, and \$2.50 per million output tokens.

\paragraph{Dataset}
We evaluate systems using the five scenarios of SemBench. Tables for each scenario are stored in a local DuckDB database \cite{raasveldt2019duckdb}, which serves as the backend for both ground-truth definitions and system predictions. We use scale factors identical to those of \citet{lao2025sembench} to generate data for each scenario.

\begin{table}
  \centering
  \begin{tabular}{l r l}
    \toprule
    \textbf{Scenario} & \textbf{Scale Factor Used} & \textbf{Modalities} \\
    \midrule
    Movie       & 2,000 & Text \textemoji   \\
    Wildlife    & 200 & Text \textemoji, Image \imageemoji, Audio \audioemoji      \\
    E-Commerce  & 500  & Text \textemoji, Image \imageemoji      \\
    Cars        & 19,672 & Text \textemoji, Image \imageemoji, Audio \audioemoji  \\
    MMQA        & 200  & Text \textemoji, Image \imageemoji       \\
    \bottomrule
  \end{tabular}
  \caption{Overview of SemBench scenarios.}
  \label{tab:sembench-scenarios}
\end{table}

\paragraph{Metrics}
We adopt the evaluation methodology of \citet{lao2025sembench} and report F1 score for retrieval queries, Spearman's rank correlation for ranking queries, and absolute error for aggregation queries. All metrics are normalized to the range [0, 1] and reported as ``quality''. 

\subsection{Baseline Systems}\label{sec:baseline-systems}

We describe the LM-DB systems evaluated in \citet{lao2025sembench} below.

\paragraph{LOTUS}

\textsc{LOTUS} \cite{patel2024semantic} extends the pandas API \cite{reback2020pandas} with additional LM-based functions. As a result, it adopts the eager execution of pandas, and therefore executes operators in the order specified by the user, without reordering. Some functions, such as the \texttt{sem\_filter} function, can be equipped with a smaller proxy model to apply a semantic cascade filter, where a decision threshold is learned at inference time and optionally used to bypass calls to the larger ``main'' LM. Other local optimizations are implemented within the specific LM functions, described in \citet{patel2024semantic}. All global logical plan optimizations (such as predicate push-down, join re-ordering, early exiting, etc.) are up to the user to implement in Python code. Given our memory-constrained local setup, we evaluate \textsc{LOTUS} identically to other systems, restricting each system to a single ``main'' LM.

\paragraph{Palimpzest}
\textsc{Palimpzest} \cite{liu2025palimpzest} is a declarative system for optimizing AI workloads over hybrid structured/unstructured data. It exposes a Python API based on lazy evaluation: users define a sequence of mixed LM/non-LM operators, which is optimized and executed upon a final \inlinesql{.run()} invocation. Though we do not evaluate it in this study, recent work such as \textsc{ABACUS} \cite{russo2025abacus}, builds on this declarative system, exploring optimization objectives when given access to LM APIs with different capability/cost trade-offs.

\paragraph{ThalamusDB}
ThalamusDB \cite{jo2024thalamusdb} is a deterministic approximate query processing system. It implements traditional relational optimizations such as single-table predicate pushdown to select table subsets in order to minimize approximation error. Additionally, it is capable of early-exiting LM functions when a \inlinesql{LIMIT} condition is satisfied. Its optimizer is unique in its multi-objective optimization over approximation error, resource usage (execution time, LLM calls, etc.), and number of labeled examples\footnote{We do not explore the labeled examples component of ThalamusDB in the present study.}.

\paragraph{BigQuery}

Released in November 2025, Google BigQuery AI functions \cite{googlebigqueryaifunctions} enable integration of LM calls into standard SQL query flows. The implementation and hyperparameters of LM inference (max concurrency, temperature, etc.) are hidden from users. Because the implementation is proprietary, we treat it as a black-box baseline.

%% file: sections/4_results.tex
\section{Results}

\subsection{Comparison to Other LM-DB Systems}\label{sec:comparison_to_other_systems}

\begin{table}[h]
\centering
\begin{tabular}{@{}lccccc@{}}
\toprule
\textbf{System} & 
\makecell{\textbf{EE}} & 
\makecell{\textbf{CF}} & 
\makecell{\textbf{CD}} & 
\makecell{\textbf{ED}} & 
\makecell{\textbf{SemBench} \\ \textbf{Coverage}} \\ \midrule
\textsc{ThalamusDB} & \textcolor{green!50!black}{\checkmark}  & \textcolor{red!80!black}{$\times$}  & \textcolor{red!80!black}{$\times$}  & \textcolor{red!80!black}{$\times$}  & 65.45\%  \\
\textsc{LOTUS}      & \textcolor{red!80!black}{$\times$}  &  \textcolor{red!80!black}{$\times$} & \textcolor{red!80!black}{$\times$}  & \textcolor{red!80!black}{$\times$}  & 78.18\%  \\
\textsc{BigQuery}   & \textcolor{blue!80!black}{\textbf{?}} & \textcolor{blue!80!black}{\textbf{?}} & \textcolor{blue!80!black}{\textbf{?}}  & \textcolor{blue!80!black}{\textbf{?}}  & 94.55\%  \\
\textsc{Palimpzest} & \textcolor{green!50!black}{\checkmark}  & \textcolor{green!50!black}{\checkmark}  & \textcolor{red!80!black}{$\times$}  & \textcolor{red!80!black}{$\times$}  & 98.18\%  \\
\textsc{BlendSQL}   & \textcolor{green!50!black}{\checkmark}  & \textcolor{green!50!black}{\checkmark}  & \textcolor{green!50!black}{\checkmark}  & \textcolor{green!50!black}{\checkmark}  & 100\%    \\ \bottomrule
\end{tabular}
\caption{System comparison on SemBench coverage and supported features. Features evaluated include early exiting (EE), cascade filtering (CF), constrained decoding (CD), and early deduplication (ED) across 55 questions.}
\label{tab:query_coverage}
\end{table}

\begin{figure}[h]
    \centering
     \includegraphics[scale=0.5]{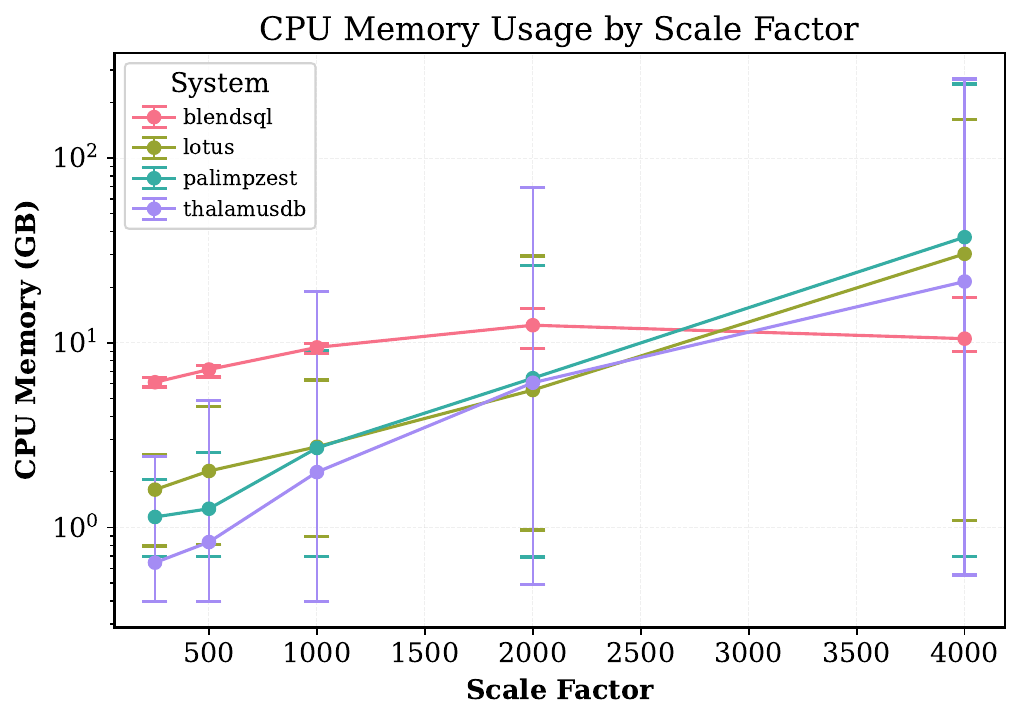}
    \caption{Plotting the average, minimum, and maximum CPU memory usage across different scale factors in the \textsc{ecomm} scenario. \textsc{BlendSQL} memory accounts for a local vLLM server running Gemma 4 E4B in addition to normal LM-DB system memory; all other systems use a remote Gemini 2.5 Flash endpoint and only account for memory usage induced by LM-DB system processing.}
    \label{fig:scalability_memory_plot}
\end{figure}

\paragraph{Query Coverage}

As shown in Table \ref{tab:query_coverage}, the design of LM-DB systems may restrict their expressivity, inhibiting full coverage across the diverse SemBench dataset. For example, \textsc{ThalamusDB} does not support LM aggregation operations, \textsc{Palimpzest} does not support top-k reranking, and \textsc{LOTUS} does not support audio inputs to LM functions. By using the two expressive polymorphic LM functions described in Section \ref{sec:lm_functions}, \textsc{BlendSQL} is the only system to display 100\% coverage over all 55 SemBench queries.

\paragraph{Benchmarking on the \textsc{movie} Scenario}
We benchmark the performance of \textsc{BlendSQL} against other open-source LM-DB systems in a 16GB VRAM setting. For all systems, we provide access to exactly one open-weight language model. While \textsc{Palimpzest} and \textsc{LOTUS} offer additional optimizations for multi-model settings, this would exceed our memory budget and is thus outside the scope of the current study. Both systems support the widest coverage of SemBench scenarios; however, we encountered unresolved compatibility issues when running either system on non-text inputs with a local \texttt{vLLM} backend, limiting our cross-system comparison to the text-only \textsc{movie} scenario.

Figure \ref{fig:comparison_against_other_systems} shows results on the text-only \textsc{movie} scenario of SemBench. Since many queries in this scenario involve \inlinesql{LIMIT} clauses (e.g., ``Find five clearly positive reviews''), the ability to early exit LM functions once a condition is satisfied is critical for achieving reasonable latency. The performance of \textsc{LOTUS} illustrates the cost of eager execution in a hybrid space of relational operators and LM functions: since the stateless LM functions have no knowledge of a future \inlinesql{LIMIT} operation, the entire subset of data must be passed to the LM function prior to executing \texttt{.head(n)}, leading to significantly higher latency. Despite this high latency, \textsc{LOTUS} yields slightly higher average quality than ThalamusDB across the four models.

Across all models, \textsc{BlendSQL} exhibits the lowest latency. It matches the quality of \textsc{Palimpzest} across Gemma 3 4B and Gemma 4 E4B, and achieves the highest quality of all systems with Gemma 3 12B. Importantly, small variations in reported quality can largely be attributed to minor prompting variations and the nondeterminism of asynchronous generation requests: as all systems except for \textsc{LOTUS} implement early exiting, final results are a product of which generation requests finish first to satisfy the exit condition.

\paragraph{Scalability to Large Databases}

Following \citet{lao2025sembench}, we perform a study on the scalability of \textsc{BlendSQL} across various database scales. We plot the CPU memory usage across \textsc{ecomm} scale factors of 250, 500, 1,000, 2,000, and 4,000 for the three baseline open-source LM-DB systems and \textsc{BlendSQL} in Figure \ref{fig:scalability_memory_plot}. While the three baseline systems' memory usage scales linearly with database size, \textsc{BlendSQL}'s usage is relatively constant, impacted only by variations in vLLM server usage. This is due to the design of the baseline systems' semantic join function: in queries such as Q8 and Q9, image pairs are constructed and evaluated against a language model without releasing the memory used to store the image pairs. As a result, each system can potentially load 25,000 .jpg images into memory at once under a scale factor of 4,000.
The three open-source systems \textsc{ThalamusDB}, \textsc{Palimpzest}, and \textsc{LOTUS} each reach a peak usage of 268GB, 252GB, and 161GB respectively under this scale factor of 4,000. Comparatively, \textsc{BlendSQL} uses a maximum of only 18GB at this scale factor to both serve the local language model with vLLM and execute the underlying multi-modal LM-DB system. This memory usage is primarily attributable to components such as the vLLM KV cache and scheduler management under high concurrency workloads: for reference, starting an idle Gemma 4 E4B vLLM server alone consumes 5GB of CPU memory.

% --- First 4 Tables (Spans both columns) ---
\begin{table*}[!t]
\centering
\caption{Latency, quality, and cost across various SemBench scenarios. All queries are clickable hyperlinks directing to the corresponding \textsc{BlendSQL} query. Green highlights indicate a better score compared to the other system. Yellow highlights indicate a tie. Quality differences within the range $\pm$ 0.02 are considered insignificant and labeled as ties with a yellow highlight. The \textsc{BlendSQL} setting is run with a max concurrency of 64.} % High-level caption
\vspace{-5pt}
\label{tab:overall_results}
\vspace{1em}

\begin{minipage}[t]{0.48\textwidth}
  \centering
  \subcaption{\textsc{mmqa} results.}
  \vspace{-0.3em} 
  \resizebox{\linewidth}{!}{\input{tables/mmqa}}
  \label{tab:mmqa}
\end{minipage}%
\hfill
\begin{minipage}[t]{0.48\textwidth}
  \centering
  \subcaption{\textsc{ecomm} results.}
  \vspace{-0.3em}
  \resizebox{\linewidth}{!}{\input{tables/ecomm}}
  \label{tab:ecomm}
\end{minipage}

% \vspace{2em} 

\begin{minipage}[t]{0.48\textwidth}
  \centering
  \subcaption{\textsc{movie} results.}
  \vspace{-0.3em}
  \resizebox{\linewidth}{!}{\input{tables/movie}}
  \label{tab:movie}
\end{minipage}%
\hfill
\begin{minipage}[t]{0.48\textwidth}
  \centering
  \subcaption{\textsc{cars} results.}
  \vspace{-0.3em}
  \resizebox{\linewidth}{!}{\input{tables/cars}}
  \label{tab:cars}
\end{minipage}
\end{table*}

% --- 5th Table (Spans only the left column, text flows on the right) ---
\begin{table}[!t]
  \ContinuedFloat % Tells LaTeX this is part of the previous table
  \centering
  \subcaption{\textsc{wildlife} results.}
  \vspace{-0.3em}
  \resizebox{\linewidth}{!}{\input{tables/wildlife}}
  \label{tab:wildlife}
\end{table}

\begin{figure*}[h]
    \centering
     \includegraphics[scale=0.5]{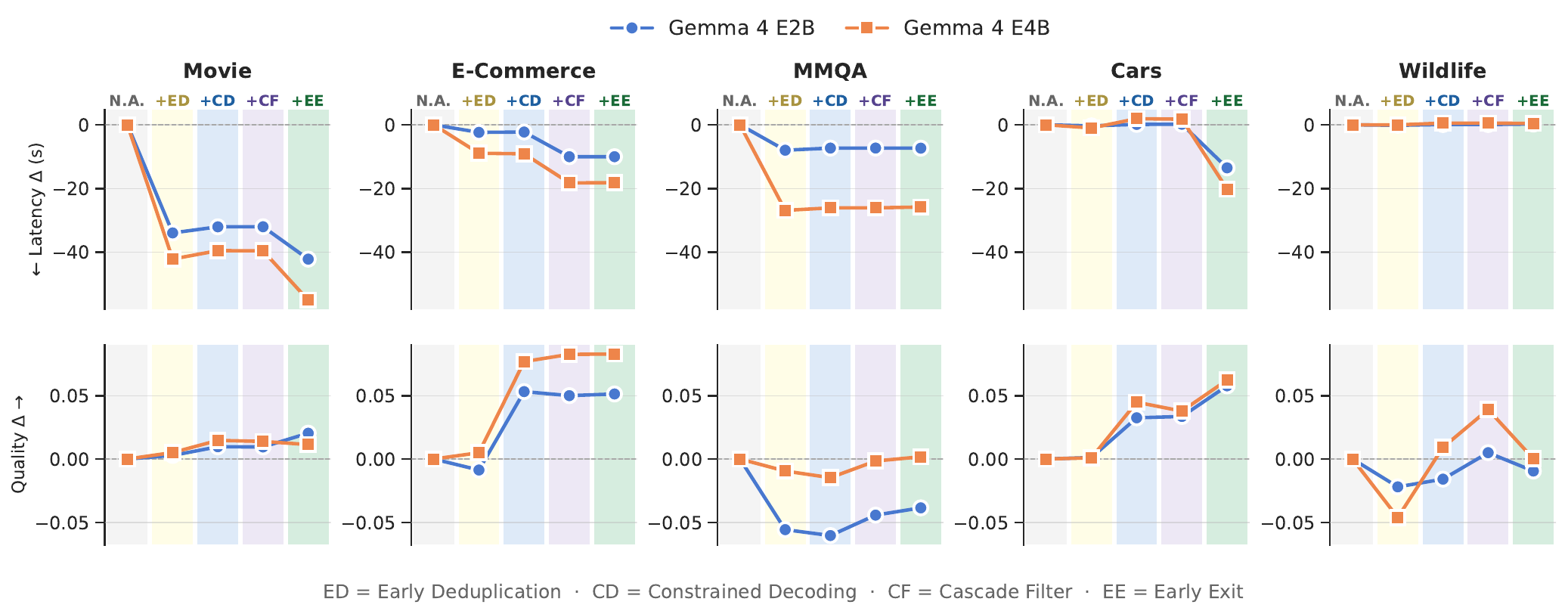}
    \caption{Measuring the relative quality and latency change when incrementally adding query processing features to \textsc{BlendSQL}. Results are averaged across 5 runs and use a max concurrency of 64. ``N.A.'' refers to v0.0.0 presented in \citet{glennblendsql}; all other features are new additions in v0.1.0.}
    \label{fig:feature_ablation_plot}
\end{figure*}

\begin{figure*}[h]
    \centering
     \includegraphics[scale=0.5]{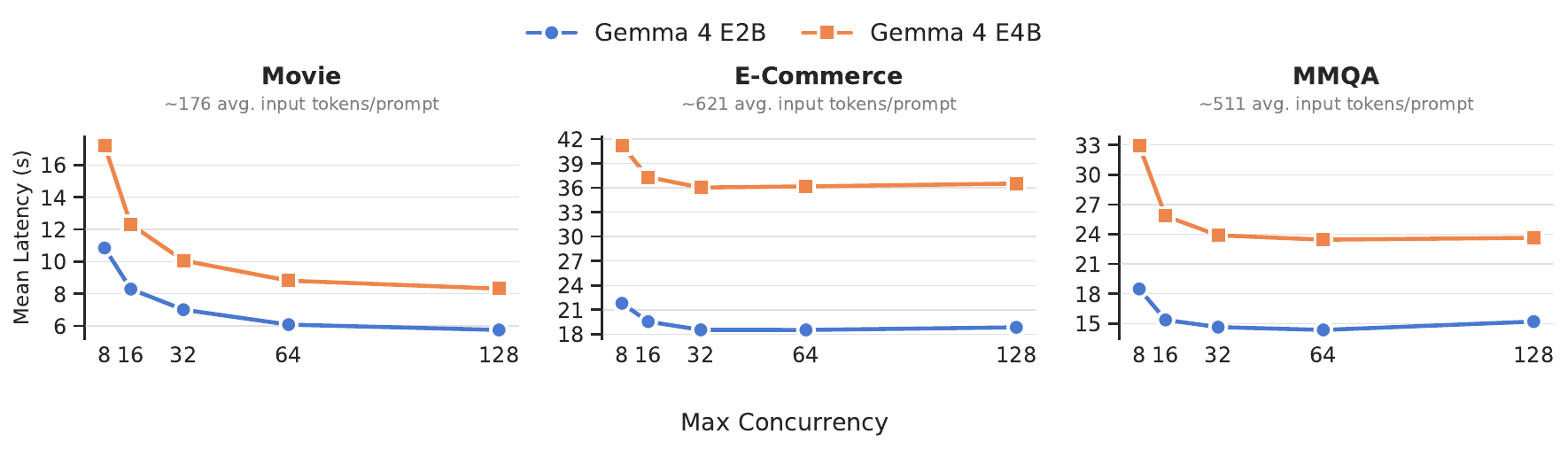}
    \caption{Evaluating the impact of concurrency limits with LM inference requests on a single 16GB RTX 5080. This controls the maximum number of concurrent async requests sent to vLLM at once, which in turn make use of continuous batching to keep GPU usage high.}
    \label{fig:concurrency_plot}
\end{figure*}

This relatively constant memory usage is made possible by the mechanism \textsc{BlendSQL} uses to achieve joins via the asynchronous \textsc{llmmap} function. Given the abbreviated \textsc{ecomm} Q8 program below, there are at most \texttt{max\_concurrency}
× 2 images loaded into memory at a given time.

\begin{lstlisting}[language=SQL,breaklines=true,showstringspaces=false]
WITH image_pairs AS (
    SELECT 
    s1.id AS id1,
    img1.local_image_path AS image1,
    s2.id AS id2,
    img2.local_image_path AS image2
    FROM styles_details s1
    JOIN image_mapping img1 ON s1.id = img1.id
    JOIN styles_details s2 ON s1.id < s2.id
    JOIN image_mapping img2 ON s2.id = img2.id
) 
SELECT id1 || '-' || id2 AS id
FROM image_pairs
WHERE {{
    LLMMap(
        'Do both images display objects of 
        the same category?',
        image1,
        image2
    )
}}
\end{lstlisting}

Running all 14 \textsc{ecomm} queries five times under the five scale factors with Gemma 4 E4B and \textsc{BlendSQL} took 38.6 hours and cost \$6.96. A direct comparison to other LM-DB systems using Gemini 2.5 Flash is difficult, since as reported in \citet{lao2025sembench}, no open-source system successfully executed all 70 queries. The closest system, \textsc{Palimpzest}, completed a single run of 65 queries in 18.6 hours and \$597.67, translating to a five-run reproduction cost of 92.8 hours and \$2,988.32.
% The above program highlights an important design philosophy of \textsc{BlendSQL}: rather than bringing database values into working memory and constructing join pairs via Python code, we can rely on the optimized execution of the native DBMS system to execute the \inlinesql{JOIN} before applying the LM function.

\subsection{Cost, Latency, and Quality of Open-Weights Models}

% We implement the 55 SemBench queries as \textsc{BlendSQL} programs and compare the cost, quality, and latency of the quantized Gemma 4 E2B and Gemma 4 E4B models to the results with Gemini 2.5 Flash \cite{comanici2025gemini} reported in \citet{lao2025sembench}. 

Table \ref{tab:overall_results} contains the cost, latency, and quality results for the evaluated SemBench scenarios across two settings: \textsc{BlendSQL} with Gemma 4 E4B, and the average SemBench system with Gemini 2.5 Flash. We define the ``average'' system as the query-level mean of the four baseline systems described in Section \ref{sec:baseline-systems}; full system-level results can be found in Appendix \ref{tab:full_sembench_results}. Visualized in Figure \ref{fig:overall_summary}, the properly harnessed small, open-weight model is competitive with the closed Gemini 2.5 Flash at a fraction of the cost: the \textsc{BlendSQL} setting demonstrates a query-level win-or-tie rate of 57\% across all scenarios and decreases experiment costs from \$226.64 to just \$0.58. Notably, the small-model \textsc{BlendSQL} setting outperforms the average SemBench system in average quality on the \textsc{mmqa} scenario (0.85 vs. 0.70) and the \textsc{wildlife} scenario (0.54 vs. 0.51). The \textsc{movie} scenario shows the \textsc{BlendSQL} system nearly matching the average SemBench system (0.72 vs. 0.74) while decreasing cost from \$25.54 to \$0.04.

\begin{figure}[h]
    \centering
     \includegraphics[scale=0.5]{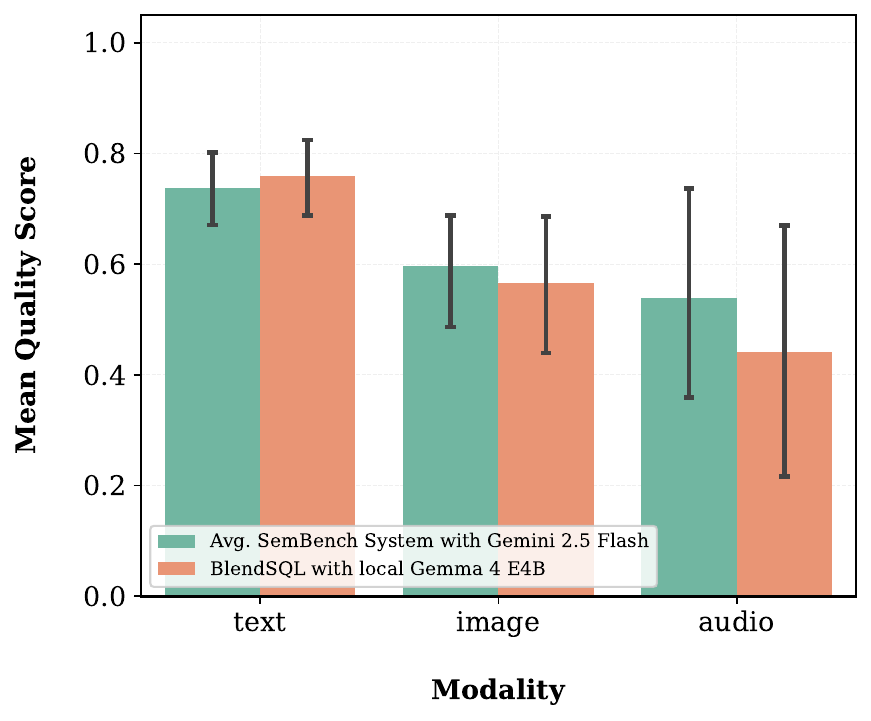}
    \caption{While a properly harnessed quantized Gemma 4 E4B slightly outperforms the closed Gemini 2.5 Flash on queries involving text, the quality gap widens on image and audio inputs, where the open-weight model exhibits a gap of 0.03 for images and 0.10 for audio.}
    \label{fig:modality_quality}
\end{figure}

We see a quality gap emerge in the audio-heavy \textsc{cars} scenario, where average quality of the \textsc{BlendSQL} setting is 0.53, falling behind the average SemBench system's 0.67. As shown in Figure \ref{fig:modality_quality}, this gap is attributable to the modality of the data being passed to the LM functions: while the small, properly harnessed Gemma E4B model outperforms the closed Gemini 2.5 Flash across text data, it falls behind on audio and image data. In Q5 of \textsc{cars}, for instance, \textsc{BlendSQL} with Gemma 4 E4B achieves a quality of 0.0: the \textsc{llmmap} function tasked with identifying recordings of damaged cars returns \texttt{FALSE} for every row, disqualifying all candidates and producing an empty result. This pattern recurs in Q2, where the model fails to identify any audio file as resembling a car with a dead battery.

\paragraph{Impact of Concurrency Limits}
% \red{TODO: is this first paragraph too 'poetic'?}
% Unburdened by provider-enforced rate limiting, open-weight models on local hardware allow for massive parallelization of requests restricted only by compute and inference software. This concurrency is critical in meeting the aggressive latency expectations of database users. Additionally, by switching to an on-demand instance pricing model rather than per-token billing, cost is tied directly to usage time rather than query volume. The adage ''time is money'' holds true: Latency optimization yields not just performance gains, but direct cost savings.

Local deployment eliminates provider-enforced rate limits, allowing request parallelism bounded only by available compute. Figure \ref{fig:concurrency_plot} displays the relationship between concurrency limits and system latency across the scenarios of SemBench.\footnote{As \texttt{vLLM} 0.21.0 does not currently support continuous batching for Gemma 4 models with audio data (\url{https://github.com/vllm-project/vllm/pull/39459}), we refrain from evaluating the two scenarios with audio data.} The text-only \textsc{movie} scenario sees the greatest speed-up with increased concurrency limits, where the Gemma 4 E4B model displays a latency drop from 17.2s to 8.3s when scaling concurrency from 8 to 128. As each \textsc{llmmap} prompt is relatively small (176 tokens on average), even just 16GB VRAM allows for the headroom required to speed up inference at large concurrency settings. By contrast, we see that the \textsc{ecomm} scenario's latency flattens at a concurrency of 32 due to the long-context prompts formed via the lengthy product descriptions (621 tokens on average). 
% ~Operating on hourly-billed cost models, the customizing of hyperparameters such as max concurrency can directly decrease latency, and therefore cost.

\paragraph{Feature Ablations}

We plot the relative impact of the query optimizations described in Section \ref{sec:query_optimization} in Figure \ref{fig:feature_ablation_plot}\footnote{Complete query-level results can be found in Appendix \ref{fig:ablation_heatmap}}. The \inlinesql{LIMIT}-heavy \textsc{movie} scenario sees a large improvement in latency when enabling early exiting, where a Gemma E4B model drops from 24.3s to 8.8s average latency. The largest change in latency is seen in the \textsc{cars} scenario, driven by two questions: In Q3, instead of performing inference over all 1,270 rows of data where \inlinesql{transmission = 'Manual'}, early exiting allows the process to complete once 10 outputs of the LM function satisfy the \inlinesql{= FALSE} predicate, resulting in latency savings of 55 seconds. In Q8, rather than calling the \textsc{llmmap} function over all 19,672 rows, processing can terminate once 100 images of cars with both punctures and paint scratches are found, saving 141 seconds.

Early deduplication is responsible for the single largest drop in latency across the dataset, saving 40 seconds on the \textsc{Movie} scenario. The reason for this is the relatively large scale factor (2,000) and level of duplication within the columns passed to LM functions: of the 2,000 values in the \inlinesql{reviewText} column, only 1,864 are distinct. This saves 136 calls to an LM on a simple \textsc{llmmap} across the column, but the effect is compounded when performing joins: the self-join logic in Q5, Q6, and Q7 decreases the size of the resulting data subset from 122,004 rows to 72,675. We see latency gains in the \textsc{mmqa} scenario for the same reasons, where Q2a and Q2b select the 6 distinct \inlinesql{Track} values rather than the 13 total values (with duplicates), resulting in the cross-join passing 1,200 values to the LM instead of 2,600.

Constrained decoding is a major factor in the quality of the language model outputs. In the \textsc{cars} scenario, the average quality scores are boosted by 0.03 and 0.04 for Gemma 4 E2B and Gemma 4 E4B, respectively. The largest increase in quality for the \textsc{cars} scenario occurs in Q10, in which the LM function must classify a customer complaint into one of 24 possible categories (electrical system, power train, engine, etc.). Applying constrained decoding results in a +0.35 improvement in quality for Gemma 4 E4B, showing that despite being prompted to return a response within a valid set of options, small language models often fail to generate a valid selection without formal runtime constraints. The largest change in quality is seen in the \textsc{ecomm} scenario, where the addition of constrained decoding boosts average quality by 0.06 for Gemma 4 E2B and 0.07 for Gemma 4 E4B. This large boost in performance is primarily due to Q12 (+0.90 increase after applying constrained decoding), which tasks the language model with generating a JSON object with a specified schema, and Q5 / Q11 (+0.05, +0.04), in which the language model plays the role of a 5-way multi-class classifier. Critically, this quality improvement is achieved without sacrificing latency: enabling constrained decoding via \textsc{llguidance} adds little-to-no overhead, adding 0.064 seconds of latency to the Gemma 4 E2B model and saving 0.20 seconds with the Gemma 4 E4B model. 

As very few queries in the SemBench benchmark contain multiple LM functions within the same subquery, cascade filtering shows little impact within this context. However, we see Q14 in the \textsc{ecomm} scenario save 142 seconds of latency after enabling cascade filtering. In this query, a program is written to (1) Find all images containing white socks, and (2) Identify which image and description pairs are aligned. By filtering on the ''white socks`` criteria first, the second function only receives the 426 filtered inputs rather than the full set of 3,000 image / description pairs.
% The reason this is possible is the mechanism that the constrained decoding engine \textsc{llguidance} \cite{guidance} uses for constructing and applying masks. Whereas some locally constrained decoding engines eagerly compile a graph of all possible token steps prior to generation, the Guidance engine constructs masks ‘on-the-fly’, meaning that the grammar compilation can be done on the otherwise idle CPU in parallel with the GPU-based generation. In this way, as long as mask creation happens faster than the language model forward pass, constrained generation will be just as fast as the unconstrained one \cite{geng2025jsonschemabench} \footnote{\url{https://guidance-ai.github.io/llguidance/llg-go-brrr}}.
% Interestingly, this scenario is unique in that both the E2B and E4B model achieve very similar quality scores (+/- 0.01) after applying constrained decoding. 

\section{Environmental Impact}

The total runtime for all reported experimental results amounted to 81 hours on the workstation described in Section \ref{sec:model_inference}. To estimate the carbon footprint of our evaluation, we followed the widely adopted methodology proposed by \citet{lannelongue2021green}\footnote{\url{https://calculator.green-algorithms.org/}}. This amounts to an estimated energy consumption of 27.21 kWh, which translates to 11.4 kgCO$_2$. For context, this carbon footprint is roughly equivalent to the emissions of driving 65.35 km in an average passenger car.

\section{Limitations}

Described in Section \ref{sec:comparison_to_other_systems}, a full comparison of \textsc{BlendSQL} against the baseline LM-DB systems with Gemma 4 E4B is not possible, as the baseline systems raise errors when using a local vLLM endpoint on multi-modal inputs. We hope that in the future, a complete comparison on the SemBench dataset with local open-weight models may be made possible.

The notion of ``financial cost'' is highly volatile and subject to the volatility of hardware rental and proprietary API pricing. The rates used in this study (\$0.18/hour for an NVIDIA RTX 5080 and \$0.30 per million input tokens for Gemini 2.5 Flash) reflect market conditions at the time of writing, and are influenced by economic factors outside the scope of this study. While monetary costs will inevitably fluctuate, our broader conclusion relies not on specific price points, but on the structural shift from a volume-based (per-token) to a time-based (hourly compute) pricing paradigm. Because \textsc{BlendSQL} v0.1.0 reduces execution time while matching or improving accuracy, this structural advantage remains robust regardless of future economic shifts. To emphasize the financial cost savings independent of on-demand rental availability in third-party GPU marketplaces, we present an alternative setting. The RTX 5080 has an MSRP of \$999.99; with standard markup pricing for high-demand graphics cards, it is available for \$1,300 as of May 2026. The AMD Ryzen 7600X is available for \$167. Once these hardware costs are accounted for, only energy costs must be considered: Electric Choice report average U.S. residential energy costs of 0.18 / kWh \cite{electricitycosts}. Factoring in these variables, the cost to buy the workstation described in Section \ref{sec:model_inference} and run the \textsc{ecomm} scalability experiment in Section \ref{sec:comparison_to_other_systems} is approximately \$1,469 (derived via $1300 + 167 + (13.13 * 0.18))$). This is still substantially less than the \$2,988 cost of using \textsc{Palimpzest} with Gemini 2.5 Flash for the same experiment.

%% file: tables/mmqa.tex
\footnotesize
\label{tab:blendsql_vs_baseline_gemma_e4b}
\begin{tabular}{l rrr | rrr}
\toprule
& \multicolumn{3}{c|}{\makecell{\textbf{BlendSQL with} \\ \texttt{gemma-4-E4B-it-FP8}}} & \multicolumn{3}{c}{\makecell{\textbf{Avg. SemBench System with} \\ \texttt{gemini-2.5-flash}}} \\
\cmidrule(lr){2-4} \cmidrule(l){5-7}
\textbf{Query} & \textbf{Lat.\,(s)} & \textbf{Quality} & \textbf{Cost\,(5\,runs)} & \textbf{Lat.\,(s)} & \textbf{Quality} & \textbf{Cost\,(5\,runs)} \\
\midrule
\href{https://github.com/CapitalOne-Research/play-by-the-type-rules/blob/main/sembench/src/queries/mmqa/blendsql/Q1.sql}{Q1 \textemoji} & \tikz[baseline=(X.base)]{\node[fill=bestgreen,rounded corners=3pt,inner xsep=3pt,inner ysep=1.5pt](X){\bfseries 0.07};} & \tikz[baseline=(X.base)]{\node[fill=tiecolor,rounded corners=3pt,inner xsep=3pt,inner ysep=1.5pt](X){\bfseries 1.00};} & \tikz[baseline=(X.base)]{\node[fill=bestgreen,rounded corners=3pt,inner xsep=3pt,inner ysep=1.5pt](X){\bfseries \$2e-5};} & 8.67 & \tikz[baseline=(X.base)]{\node[fill=tiecolor,rounded corners=3pt,inner xsep=3pt,inner ysep=1.5pt](X){\bfseries 1.00};} & \$0.06 \\
\rowcolor{rowgray}
\href{https://github.com/CapitalOne-Research/play-by-the-type-rules/blob/main/sembench/src/queries/mmqa/blendsql/Q2a.sql}{Q2a \textemoji \imageemoji} & \tikz[baseline=(X.base)]{\node[fill=bestgreen,rounded corners=3pt,inner xsep=3pt,inner ysep=1.5pt](X){\bfseries 55.15};} & \tikz[baseline=(X.base)]{\node[fill=bestgreen,rounded corners=3pt,inner xsep=3pt,inner ysep=1.5pt](X){\bfseries 1.00};} & \tikz[baseline=(X.base)]{\node[fill=bestgreen,rounded corners=3pt,inner xsep=3pt,inner ysep=1.5pt](X){\bfseries \$0.01};} & 119.77 & 0.61 & \$3.35 \\
\href{https://github.com/CapitalOne-Research/play-by-the-type-rules/blob/main/sembench/src/queries/mmqa/blendsql/Q2b.sql}{Q2b \textemoji \imageemoji} & \tikz[baseline=(X.base)]{\node[fill=bestgreen,rounded corners=3pt,inner xsep=3pt,inner ysep=1.5pt](X){\bfseries 64.22};} & \tikz[baseline=(X.base)]{\node[fill=bestgreen,rounded corners=3pt,inner xsep=3pt,inner ysep=1.5pt](X){\bfseries 0.97};} & \tikz[baseline=(X.base)]{\node[fill=bestgreen,rounded corners=3pt,inner xsep=3pt,inner ysep=1.5pt](X){\bfseries \$0.02};} & 113.00 & 0.61 & \$3.42 \\
\rowcolor{rowgray}
Q3a & \tikz[baseline=(X.base)]{\node[fill=bestgreen,rounded corners=3pt,inner xsep=3pt,inner ysep=1.5pt](X){\bfseries 1.35};} & 0.78 & \tikz[baseline=(X.base)]{\node[fill=bestgreen,rounded corners=3pt,inner xsep=3pt,inner ysep=1.5pt](X){\bfseries \$3e-4};} & 9.82 & \tikz[baseline=(X.base)]{\node[fill=bestgreen,rounded corners=3pt,inner xsep=3pt,inner ysep=1.5pt](X){\bfseries 0.82};} & \$1.34 \\
\href{https://github.com/CapitalOne-Research/play-by-the-type-rules/blob/main/sembench/src/queries/mmqa/blendsql/Q3f.sql}{Q3f \textemoji} & \tikz[baseline=(X.base)]{\node[fill=bestgreen,rounded corners=3pt,inner xsep=3pt,inner ysep=1.5pt](X){\bfseries 1.41};} & \tikz[baseline=(X.base)]{\node[fill=tiecolor,rounded corners=3pt,inner xsep=3pt,inner ysep=1.5pt](X){\bfseries 0.92};} & \tikz[baseline=(X.base)]{\node[fill=bestgreen,rounded corners=3pt,inner xsep=3pt,inner ysep=1.5pt](X){\bfseries \$4e-4};} & 10.38 & \tikz[baseline=(X.base)]{\node[fill=tiecolor,rounded corners=3pt,inner xsep=3pt,inner ysep=1.5pt](X){\bfseries 0.92};} & \$0.06 \\
\rowcolor{rowgray}
\href{https://github.com/CapitalOne-Research/play-by-the-type-rules/blob/main/sembench/src/queries/mmqa/blendsql/Q4.sql}{Q4 \textemoji} & \tikz[baseline=(X.base)]{\node[fill=bestgreen,rounded corners=3pt,inner xsep=3pt,inner ysep=1.5pt](X){\bfseries 0.45};} & \tikz[baseline=(X.base)]{\node[fill=tiecolor,rounded corners=3pt,inner xsep=3pt,inner ysep=1.5pt](X){\bfseries 0.56};} & \tikz[baseline=(X.base)]{\node[fill=bestgreen,rounded corners=3pt,inner xsep=3pt,inner ysep=1.5pt](X){\bfseries \$1e-4};} & 1.20 & \tikz[baseline=(X.base)]{\node[fill=tiecolor,rounded corners=3pt,inner xsep=3pt,inner ysep=1.5pt](X){\bfseries 0.54};} & \$0.03 \\
\href{https://github.com/CapitalOne-Research/play-by-the-type-rules/blob/main/sembench/src/queries/mmqa/blendsql/Q5.sql}{Q5 \textemoji} & \tikz[baseline=(X.base)]{\node[fill=bestgreen,rounded corners=3pt,inner xsep=3pt,inner ysep=1.5pt](X){\bfseries 0.24};} & \tikz[baseline=(X.base)]{\node[fill=tiecolor,rounded corners=3pt,inner xsep=3pt,inner ysep=1.5pt](X){\bfseries 1.00};} & \tikz[baseline=(X.base)]{\node[fill=bestgreen,rounded corners=3pt,inner xsep=3pt,inner ysep=1.5pt](X){\bfseries \$6e-5};} & 0.49 & \tikz[baseline=(X.base)]{\node[fill=tiecolor,rounded corners=3pt,inner xsep=3pt,inner ysep=1.5pt](X){\bfseries 1.00};} & \$5e-3 \\
\rowcolor{rowgray}
\href{https://github.com/CapitalOne-Research/play-by-the-type-rules/blob/main/sembench/src/queries/mmqa/blendsql/Q6a.sql}{Q6a \textemoji} & \tikz[baseline=(X.base)]{\node[fill=bestgreen,rounded corners=3pt,inner xsep=3pt,inner ysep=1.5pt](X){\bfseries 0.73};} & \tikz[baseline=(X.base)]{\node[fill=bestgreen,rounded corners=3pt,inner xsep=3pt,inner ysep=1.5pt](X){\bfseries 1.00};} & \tikz[baseline=(X.base)]{\node[fill=bestgreen,rounded corners=3pt,inner xsep=3pt,inner ysep=1.5pt](X){\bfseries \$2e-4};} & 8.22 & 0.59 & \$0.05 \\
\href{https://github.com/CapitalOne-Research/play-by-the-type-rules/blob/main/sembench/src/queries/mmqa/blendsql/Q6b.sql}{Q6b \textemoji} & \tikz[baseline=(X.base)]{\node[fill=bestgreen,rounded corners=3pt,inner xsep=3pt,inner ysep=1.5pt](X){\bfseries 0.71};} & 0.67 & \tikz[baseline=(X.base)]{\node[fill=bestgreen,rounded corners=3pt,inner xsep=3pt,inner ysep=1.5pt](X){\bfseries \$2e-4};} & 7.67 & \tikz[baseline=(X.base)]{\node[fill=bestgreen,rounded corners=3pt,inner xsep=3pt,inner ysep=1.5pt](X){\bfseries 0.76};} & \$0.05 \\
\rowcolor{rowgray}
\href{https://github.com/CapitalOne-Research/play-by-the-type-rules/blob/main/sembench/src/queries/mmqa/blendsql/Q6c.sql}{Q6c \textemoji} & \tikz[baseline=(X.base)]{\node[fill=bestgreen,rounded corners=3pt,inner xsep=3pt,inner ysep=1.5pt](X){\bfseries 0.83};} & \tikz[baseline=(X.base)]{\node[fill=bestgreen,rounded corners=3pt,inner xsep=3pt,inner ysep=1.5pt](X){\bfseries 0.91};} & \tikz[baseline=(X.base)]{\node[fill=bestgreen,rounded corners=3pt,inner xsep=3pt,inner ysep=1.5pt](X){\bfseries \$2e-4};} & 10.02 & 0.67 & \$0.05 \\
\href{https://github.com/CapitalOne-Research/play-by-the-type-rules/blob/main/sembench/src/queries/mmqa/blendsql/Q7.sql}{Q7 \textemoji \imageemoji} & \tikz[baseline=(X.base)]{\node[fill=bestgreen,rounded corners=3pt,inner xsep=3pt,inner ysep=1.5pt](X){\bfseries 133.26};} & \tikz[baseline=(X.base)]{\node[fill=bestgreen,rounded corners=3pt,inner xsep=3pt,inner ysep=1.5pt](X){\bfseries 0.51};} & \tikz[baseline=(X.base)]{\node[fill=bestgreen,rounded corners=3pt,inner xsep=3pt,inner ysep=1.5pt](X){\bfseries \$0.03};} & 1501.57 & 0.21 & \$50.73 \\
\midrule
\textbf{Avg.} & \tikz[baseline=(X.base)]{\node[fill=bestgreen,rounded corners=3pt,inner xsep=3pt,inner ysep=1.5pt](X){\bfseries 23.49};} & \tikz[baseline=(X.base)]{\node[fill=bestgreen,rounded corners=3pt,inner xsep=3pt,inner ysep=1.5pt](X){\bfseries 0.85};} & \tikz[baseline=(X.base)]{\node[fill=bestgreen,rounded corners=3pt,inner xsep=3pt,inner ysep=1.5pt](X){\bfseries \$6e-3};} & 163.19 & 0.71 & \$5.37 \\
\textbf{Avg.\ Std.} & 0.11 & 0.03 & \$3e-5 & - & - & - \\
\textbf{Wins} & 11 & 5 & 11 & 0 & 2 & 0 \\
\textbf{Total Cost} & & & \tikz[baseline=(X.base)]{\node[fill=bestgreen,rounded corners=3pt,inner xsep=3pt,inner ysep=1.5pt](X){\bfseries \$0.06};} & & & \$59.12 \\
\bottomrule
\end{tabular}

%% file: tables/ecomm.tex
\footnotesize
\label{tab:blendsql_vs_baseline_gemma_e4b}
\begin{tabular}{l rrr | rrr}
\toprule
 & \multicolumn{3}{c|}{\makecell{\textbf{BlendSQL with} \\ \texttt{gemma-4-E4B-it-FP8}}} & \multicolumn{3}{c}{\makecell{\textbf{Avg. SemBench System with} \\ \texttt{gemini-2.5-flash}}} \\
\cmidrule(lr){2-4} \cmidrule(l){5-7}
\textbf{Query} & \textbf{Lat.\,(s)} & \textbf{Quality} & \textbf{Cost\,(5\,runs)} & \textbf{Lat.\,(s)} & \textbf{Quality} & \textbf{Cost\,(5\,runs)} \\
\midrule
\href{https://github.com/CapitalOne-Research/play-by-the-type-rules/blob/main/sembench/src/queries/ecomm/blendsql/Q1.sql}{Q1 \textemoji} & \tikz[baseline=(X.base)]{\node[fill=bestgreen,rounded corners=3pt,inner xsep=3pt,inner ysep=1.5pt](X){\bfseries 8.18};} & \tikz[baseline=(X.base)]{\node[fill=bestgreen,rounded corners=3pt,inner xsep=3pt,inner ysep=1.5pt](X){\bfseries 1.00};} & \tikz[baseline=(X.base)]{\node[fill=bestgreen,rounded corners=3pt,inner xsep=3pt,inner ysep=1.5pt](X){\bfseries \$2e-3};} & 20.10 & 0.90 & \$0.26 \\
\rowcolor{rowgray}
\href{https://github.com/CapitalOne-Research/play-by-the-type-rules/blob/main/sembench/src/queries/ecomm/blendsql/Q2.sql}{Q2 \imageemoji} & \tikz[baseline=(X.base)]{\node[fill=bestgreen,rounded corners=3pt,inner xsep=3pt,inner ysep=1.5pt](X){\bfseries 23.15};} & 0.40 & \tikz[baseline=(X.base)]{\node[fill=bestgreen,rounded corners=3pt,inner xsep=3pt,inner ysep=1.5pt](X){\bfseries \$6e-3};} & 95.03 & \tikz[baseline=(X.base)]{\node[fill=bestgreen,rounded corners=3pt,inner xsep=3pt,inner ysep=1.5pt](X){\bfseries 0.65};} & \$6.04 \\
\href{https://github.com/CapitalOne-Research/play-by-the-type-rules/blob/main/sembench/src/queries/ecomm/blendsql/Q3.sql}{Q3 \textemoji} & \tikz[baseline=(X.base)]{\node[fill=bestgreen,rounded corners=3pt,inner xsep=3pt,inner ysep=1.5pt](X){\bfseries 8.53};} & 0.78 & \tikz[baseline=(X.base)]{\node[fill=bestgreen,rounded corners=3pt,inner xsep=3pt,inner ysep=1.5pt](X){\bfseries \$2e-3};} & 18.27 & \tikz[baseline=(X.base)]{\node[fill=bestgreen,rounded corners=3pt,inner xsep=3pt,inner ysep=1.5pt](X){\bfseries 0.97};} & \$0.52 \\
\rowcolor{rowgray}
\href{https://github.com/CapitalOne-Research/play-by-the-type-rules/blob/main/sembench/src/queries/ecomm/blendsql/Q4.sql}{Q4 \imageemoji} & \tikz[baseline=(X.base)]{\node[fill=bestgreen,rounded corners=3pt,inner xsep=3pt,inner ysep=1.5pt](X){\bfseries 13.65};} & 0.40 & \tikz[baseline=(X.base)]{\node[fill=bestgreen,rounded corners=3pt,inner xsep=3pt,inner ysep=1.5pt](X){\bfseries \$3e-3};} & 174.23 & \tikz[baseline=(X.base)]{\node[fill=bestgreen,rounded corners=3pt,inner xsep=3pt,inner ysep=1.5pt](X){\bfseries 0.56};} & \$1.25 \\
\href{https://github.com/CapitalOne-Research/play-by-the-type-rules/blob/main/sembench/src/queries/ecomm/blendsql/Q5.sql}{Q5 \textemoji} & \tikz[baseline=(X.base)]{\node[fill=bestgreen,rounded corners=3pt,inner xsep=3pt,inner ysep=1.5pt](X){\bfseries 4.48};} & 0.96 & \tikz[baseline=(X.base)]{\node[fill=bestgreen,rounded corners=3pt,inner xsep=3pt,inner ysep=1.5pt](X){\bfseries \$1e-3};} & 13.23 & \tikz[baseline=(X.base)]{\node[fill=bestgreen,rounded corners=3pt,inner xsep=3pt,inner ysep=1.5pt](X){\bfseries 0.98};} & \$0.47 \\
\rowcolor{rowgray}
\href{https://github.com/CapitalOne-Research/play-by-the-type-rules/blob/main/sembench/src/queries/ecomm/blendsql/Q6.sql}{Q6 \imageemoji} & \tikz[baseline=(X.base)]{\node[fill=bestgreen,rounded corners=3pt,inner xsep=3pt,inner ysep=1.5pt](X){\bfseries 13.34};} & 0.67 & \tikz[baseline=(X.base)]{\node[fill=bestgreen,rounded corners=3pt,inner xsep=3pt,inner ysep=1.5pt](X){\bfseries \$3e-3};} & 97.80 & \tikz[baseline=(X.base)]{\node[fill=bestgreen,rounded corners=3pt,inner xsep=3pt,inner ysep=1.5pt](X){\bfseries 0.89};} & \$1.50 \\
\href{https://github.com/CapitalOne-Research/play-by-the-type-rules/blob/main/sembench/src/queries/ecomm/blendsql/Q7.sql}{Q7 \textemoji} & 222.74 & \tikz[baseline=(X.base)]{\node[fill=bestgreen,rounded corners=3pt,inner xsep=3pt,inner ysep=1.5pt](X){\bfseries 0.77};} & \tikz[baseline=(X.base)]{\node[fill=bestgreen,rounded corners=3pt,inner xsep=3pt,inner ysep=1.5pt](X){\bfseries \$0.06};} & \tikz[baseline=(X.base)]{\node[fill=bestgreen,rounded corners=3pt,inner xsep=3pt,inner ysep=1.5pt](X){\bfseries 157.53};} & 0.75 & \$5.08 \\
\rowcolor{rowgray}
\href{https://github.com/CapitalOne-Research/play-by-the-type-rules/blob/main/sembench/src/queries/ecomm/blendsql/Q8.sql}{Q8 \textemoji \imageemoji} & \tikz[baseline=(X.base)]{\node[fill=bestgreen,rounded corners=3pt,inner xsep=3pt,inner ysep=1.5pt](X){\bfseries 51.63};} & 0.50 & \tikz[baseline=(X.base)]{\node[fill=bestgreen,rounded corners=3pt,inner xsep=3pt,inner ysep=1.5pt](X){\bfseries \$0.01};} & 211.80 & \tikz[baseline=(X.base)]{\node[fill=bestgreen,rounded corners=3pt,inner xsep=3pt,inner ysep=1.5pt](X){\bfseries 0.57};} & \$23.18 \\
\href{https://github.com/CapitalOne-Research/play-by-the-type-rules/blob/main/sembench/src/queries/ecomm/blendsql/Q9.sql}{Q9 \imageemoji} & \tikz[baseline=(X.base)]{\node[fill=bestgreen,rounded corners=3pt,inner xsep=3pt,inner ysep=1.5pt](X){\bfseries 24.43};} & \tikz[baseline=(X.base)]{\node[fill=bestgreen,rounded corners=3pt,inner xsep=3pt,inner ysep=1.5pt](X){\bfseries 0.43};} & \tikz[baseline=(X.base)]{\node[fill=bestgreen,rounded corners=3pt,inner xsep=3pt,inner ysep=1.5pt](X){\bfseries \$6e-3};} & 302.27 & 0.41 & \$0.85 \\
\rowcolor{rowgray}
\href{https://github.com/CapitalOne-Research/play-by-the-type-rules/blob/main/sembench/src/queries/ecomm/blendsql/Q10.sql}{Q10 \imageemoji} & \tikz[baseline=(X.base)]{\node[fill=bestgreen,rounded corners=3pt,inner xsep=3pt,inner ysep=1.5pt](X){\bfseries 10.95};} & 0.00 & \tikz[baseline=(X.base)]{\node[fill=bestgreen,rounded corners=3pt,inner xsep=3pt,inner ysep=1.5pt](X){\bfseries \$3e-3};} & 856.00 & \tikz[baseline=(X.base)]{\node[fill=bestgreen,rounded corners=3pt,inner xsep=3pt,inner ysep=1.5pt](X){\bfseries 0.03};} & \$2.70 \\
\href{https://github.com/CapitalOne-Research/play-by-the-type-rules/blob/main/sembench/src/queries/ecomm/blendsql/Q11.sql}{Q11 \textemoji \imageemoji} & \tikz[baseline=(X.base)]{\node[fill=bestgreen,rounded corners=3pt,inner xsep=3pt,inner ysep=1.5pt](X){\bfseries 57.46};} & 0.25 & \tikz[baseline=(X.base)]{\node[fill=bestgreen,rounded corners=3pt,inner xsep=3pt,inner ysep=1.5pt](X){\bfseries \$0.01};} & 145.55 & \tikz[baseline=(X.base)]{\node[fill=bestgreen,rounded corners=3pt,inner xsep=3pt,inner ysep=1.5pt](X){\bfseries 0.76};} & \$2.20 \\
\rowcolor{rowgray}
\href{https://github.com/CapitalOne-Research/play-by-the-type-rules/blob/main/sembench/src/queries/ecomm/blendsql/Q12.sql}{Q12 \textemoji \imageemoji} & \tikz[baseline=(X.base)]{\node[fill=bestgreen,rounded corners=3pt,inner xsep=3pt,inner ysep=1.5pt](X){\bfseries 13.58};} & \tikz[baseline=(X.base)]{\node[fill=bestgreen,rounded corners=3pt,inner xsep=3pt,inner ysep=1.5pt](X){\bfseries 0.91};} & \tikz[baseline=(X.base)]{\node[fill=bestgreen,rounded corners=3pt,inner xsep=3pt,inner ysep=1.5pt](X){\bfseries \$3e-3};} & 33.13 & 0.52 & \$0.57 \\
\href{https://github.com/CapitalOne-Research/play-by-the-type-rules/blob/main/sembench/src/queries/ecomm/blendsql/Q13.sql}{Q13 \textemoji \imageemoji} & \tikz[baseline=(X.base)]{\node[fill=bestgreen,rounded corners=3pt,inner xsep=3pt,inner ysep=1.5pt](X){\bfseries 32.99};} & 0.51 & \tikz[baseline=(X.base)]{\node[fill=bestgreen,rounded corners=3pt,inner xsep=3pt,inner ysep=1.5pt](X){\bfseries \$8e-3};} & 178.50 & \tikz[baseline=(X.base)]{\node[fill=bestgreen,rounded corners=3pt,inner xsep=3pt,inner ysep=1.5pt](X){\bfseries 0.73};} & \$1.77 \\
\rowcolor{rowgray}
\href{https://github.com/CapitalOne-Research/play-by-the-type-rules/blob/main/sembench/src/queries/ecomm/blendsql/Q14.sql}{Q14 \textemoji \imageemoji} & \tikz[baseline=(X.base)]{\node[fill=bestgreen,rounded corners=3pt,inner xsep=3pt,inner ysep=1.5pt](X){\bfseries 48.73};} & \tikz[baseline=(X.base)]{\node[fill=bestgreen,rounded corners=3pt,inner xsep=3pt,inner ysep=1.5pt](X){\bfseries 0.88};} & \tikz[baseline=(X.base)]{\node[fill=bestgreen,rounded corners=3pt,inner xsep=3pt,inner ysep=1.5pt](X){\bfseries \$0.01};} & 125.90 & 0.62 & \$11.23 \\
\midrule
\textbf{Avg.} & \tikz[baseline=(X.base)]{\node[fill=bestgreen,rounded corners=3pt,inner xsep=3pt,inner ysep=1.5pt](X){\bfseries 38.13};} & 0.60 & \tikz[baseline=(X.base)]{\node[fill=bestgreen,rounded corners=3pt,inner xsep=3pt,inner ysep=1.5pt](X){\bfseries \$1e-2};} & 173.52 & \tikz[baseline=(X.base)]{\node[fill=bestgreen,rounded corners=3pt,inner xsep=3pt,inner ysep=1.5pt](X){\bfseries 0.67};} & \$4.11 \\
\textbf{Avg.\ Std.} & 0.15 & 0.02 & \$4e-5 & - & - & - \\
\textbf{Wins} & 13 & 5 & 14 & 1 & 9 & 0 \\
\textbf{Total Cost} & & & \tikz[baseline=(X.base)]{\node[fill=bestgreen,rounded corners=3pt,inner xsep=3pt,inner ysep=1.5pt](X){\bfseries \$0.13};} & & & \$57.60 \\
\bottomrule
\end{tabular}

%% file: tables/movie.tex
\footnotesize
\label{tab:blendsql_vs_baseline_gemma_e4b}
\begin{tabular}{l rrr | rrr}
\toprule
& \multicolumn{3}{c|}{\makecell{\textbf{BlendSQL with} \\ \texttt{gemma-4-E4B-it-FP8}}} & \multicolumn{3}{c}{\makecell{\textbf{Avg. SemBench System with} \\ \texttt{gemini-2.5-flash}}} \\
\cmidrule(lr){2-4} \cmidrule(l){5-7}
\textbf{Query} & \textbf{Lat.\,(s)} & \textbf{Quality} & \textbf{Cost\,(5\,runs)} & \textbf{Lat.\,(s)} & \textbf{Quality} & \textbf{Cost\,(5\,runs)} \\
\midrule
\href{https://github.com/CapitalOne-Research/play-by-the-type-rules/blob/main/sembench/src/queries/movie/blendsql/Q1.sql}{Q1 \textemoji} & \tikz[baseline=(X.base)]{\node[fill=bestgreen,rounded corners=3pt,inner xsep=3pt,inner ysep=1.5pt](X){\bfseries 0.39};} & \tikz[baseline=(X.base)]{\node[fill=tiecolor,rounded corners=3pt,inner xsep=3pt,inner ysep=1.5pt](X){\bfseries 1.00};} & \tikz[baseline=(X.base)]{\node[fill=bestgreen,rounded corners=3pt,inner xsep=3pt,inner ysep=1.5pt](X){\bfseries \$1e-4};} & 16.85 & \tikz[baseline=(X.base)]{\node[fill=tiecolor,rounded corners=3pt,inner xsep=3pt,inner ysep=1.5pt](X){\bfseries 0.99};} & \$0.18 \\
\rowcolor{rowgray}
\href{https://github.com/CapitalOne-Research/play-by-the-type-rules/blob/main/sembench/src/queries/movie/blendsql/Q2.sql}{Q2 \textemoji} & \tikz[baseline=(X.base)]{\node[fill=bestgreen,rounded corners=3pt,inner xsep=3pt,inner ysep=1.5pt](X){\bfseries 0.45};} & 0.96 & \tikz[baseline=(X.base)]{\node[fill=bestgreen,rounded corners=3pt,inner xsep=3pt,inner ysep=1.5pt](X){\bfseries \$1e-4};} & 10.80 & \tikz[baseline=(X.base)]{\node[fill=bestgreen,rounded corners=3pt,inner xsep=3pt,inner ysep=1.5pt](X){\bfseries 0.98};} & \$0.03 \\
\href{https://github.com/CapitalOne-Research/play-by-the-type-rules/blob/main/sembench/src/queries/movie/blendsql/Q3.sql}{Q3 \textemoji} & \tikz[baseline=(X.base)]{\node[fill=bestgreen,rounded corners=3pt,inner xsep=3pt,inner ysep=1.5pt](X){\bfseries 0.48};} & \tikz[baseline=(X.base)]{\node[fill=tiecolor,rounded corners=3pt,inner xsep=3pt,inner ysep=1.5pt](X){\bfseries 0.69};} & \tikz[baseline=(X.base)]{\node[fill=bestgreen,rounded corners=3pt,inner xsep=3pt,inner ysep=1.5pt](X){\bfseries \$1e-4};} & 5.20 & \tikz[baseline=(X.base)]{\node[fill=tiecolor,rounded corners=3pt,inner xsep=3pt,inner ysep=1.5pt](X){\bfseries 0.67};} & \$0.03 \\
\rowcolor{rowgray}
\href{https://github.com/CapitalOne-Research/play-by-the-type-rules/blob/main/sembench/src/queries/movie/blendsql/Q4.sql}{Q4 \textemoji} & \tikz[baseline=(X.base)]{\node[fill=bestgreen,rounded corners=3pt,inner xsep=3pt,inner ysep=1.5pt](X){\bfseries 0.32};} & 0.64 & \tikz[baseline=(X.base)]{\node[fill=bestgreen,rounded corners=3pt,inner xsep=3pt,inner ysep=1.5pt](X){\bfseries \$8e-5};} & 5.60 & \tikz[baseline=(X.base)]{\node[fill=bestgreen,rounded corners=3pt,inner xsep=3pt,inner ysep=1.5pt](X){\bfseries 0.69};} & \$0.04 \\
\href{https://github.com/CapitalOne-Research/play-by-the-type-rules/blob/main/sembench/src/queries/movie/blendsql/Q5.sql}{Q5 \textemoji} & \tikz[baseline=(X.base)]{\node[fill=bestgreen,rounded corners=3pt,inner xsep=3pt,inner ysep=1.5pt](X){\bfseries 0.63};} & 0.46 & \tikz[baseline=(X.base)]{\node[fill=bestgreen,rounded corners=3pt,inner xsep=3pt,inner ysep=1.5pt](X){\bfseries \$2e-4};} & 148.80 & \tikz[baseline=(X.base)]{\node[fill=bestgreen,rounded corners=3pt,inner xsep=3pt,inner ysep=1.5pt](X){\bfseries 0.72};} & \$4.25 \\
\rowcolor{rowgray}
\href{https://github.com/CapitalOne-Research/play-by-the-type-rules/blob/main/sembench/src/queries/movie/blendsql/Q6.sql}{Q6 \textemoji} & \tikz[baseline=(X.base)]{\node[fill=bestgreen,rounded corners=3pt,inner xsep=3pt,inner ysep=1.5pt](X){\bfseries 0.57};} & 0.46 & \tikz[baseline=(X.base)]{\node[fill=bestgreen,rounded corners=3pt,inner xsep=3pt,inner ysep=1.5pt](X){\bfseries \$1e-4};} & 122.72 & \tikz[baseline=(X.base)]{\node[fill=bestgreen,rounded corners=3pt,inner xsep=3pt,inner ysep=1.5pt](X){\bfseries 0.76};} & \$3.53 \\
\href{https://github.com/CapitalOne-Research/play-by-the-type-rules/blob/main/sembench/src/queries/movie/blendsql/Q7.sql}{Q7 \textemoji} & \tikz[baseline=(X.base)]{\node[fill=bestgreen,rounded corners=3pt,inner xsep=3pt,inner ysep=1.5pt](X){\bfseries 149.03};} & \tikz[baseline=(X.base)]{\node[fill=tiecolor,rounded corners=3pt,inner xsep=3pt,inner ysep=1.5pt](X){\bfseries 0.54};} & \tikz[baseline=(X.base)]{\node[fill=bestgreen,rounded corners=3pt,inner xsep=3pt,inner ysep=1.5pt](X){\bfseries \$0.04};} & 321.00 & \tikz[baseline=(X.base)]{\node[fill=tiecolor,rounded corners=3pt,inner xsep=3pt,inner ysep=1.5pt](X){\bfseries 0.54};} & \$16.24 \\
\rowcolor{rowgray}
\href{https://github.com/CapitalOne-Research/play-by-the-type-rules/blob/main/sembench/src/queries/movie/blendsql/Q8.sql}{Q8 \textemoji} & \tikz[baseline=(X.base)]{\node[fill=bestgreen,rounded corners=3pt,inner xsep=3pt,inner ysep=1.5pt](X){\bfseries 0.68};} & \tikz[baseline=(X.base)]{\node[fill=bestgreen,rounded corners=3pt,inner xsep=3pt,inner ysep=1.5pt](X){\bfseries 0.92};} & \tikz[baseline=(X.base)]{\node[fill=bestgreen,rounded corners=3pt,inner xsep=3pt,inner ysep=1.5pt](X){\bfseries \$2e-4};} & 6.08 & 0.84 & \$0.04 \\
\href{https://github.com/CapitalOne-Research/play-by-the-type-rules/blob/main/sembench/src/queries/movie/blendsql/Q9.sql}{Q9 \textemoji} & \tikz[baseline=(X.base)]{\node[fill=bestgreen,rounded corners=3pt,inner xsep=3pt,inner ysep=1.5pt](X){\bfseries 0.81};} & \tikz[baseline=(X.base)]{\node[fill=bestgreen,rounded corners=3pt,inner xsep=3pt,inner ysep=1.5pt](X){\bfseries 0.87};} & \tikz[baseline=(X.base)]{\node[fill=bestgreen,rounded corners=3pt,inner xsep=3pt,inner ysep=1.5pt](X){\bfseries \$2e-4};} & 7.97 & 0.77 & \$0.15 \\
\rowcolor{rowgray}
\href{https://github.com/CapitalOne-Research/play-by-the-type-rules/blob/main/sembench/src/queries/movie/blendsql/Q10.sql}{Q10 \textemoji} & \tikz[baseline=(X.base)]{\node[fill=bestgreen,rounded corners=3pt,inner xsep=3pt,inner ysep=1.5pt](X){\bfseries 8.88};} & \tikz[baseline=(X.base)]{\node[fill=bestgreen,rounded corners=3pt,inner xsep=3pt,inner ysep=1.5pt](X){\bfseries 0.71};} & \tikz[baseline=(X.base)]{\node[fill=bestgreen,rounded corners=3pt,inner xsep=3pt,inner ysep=1.5pt](X){\bfseries \$2e-3};} & 34.07 & 0.42 & \$1.07 \\
\midrule
\textbf{Avg.} & \tikz[baseline=(X.base)]{\node[fill=bestgreen,rounded corners=3pt,inner xsep=3pt,inner ysep=1.5pt](X){\bfseries 16.22};} & 0.72 & \tikz[baseline=(X.base)]{\node[fill=bestgreen,rounded corners=3pt,inner xsep=3pt,inner ysep=1.5pt](X){\bfseries \$4e-3};} & 67.91 & \tikz[baseline=(X.base)]{\node[fill=bestgreen,rounded corners=3pt,inner xsep=3pt,inner ysep=1.5pt](X){\bfseries 0.74};} & \$2.55 \\
\textbf{Avg.\ Std.} & 0.22 & 0.03 & \$6e-5 & - & - & \- \\
\textbf{Wins} & 10 & 3 & 10 & 0 & 4 & 0 \\
\textbf{Total Cost} & & & \tikz[baseline=(X.base)]{\node[fill=bestgreen,rounded corners=3pt,inner xsep=3pt,inner ysep=1.5pt](X){\bfseries \$0.04};} & & & \$25.54 \\
\bottomrule
\end{tabular}

%% file: tables/cars.tex
\centering
\footnotesize
\label{tab:blendsql_vs_baseline_gemma_e4b}
\begin{tabular}{l rrr | rrr}
\toprule
 & \multicolumn{3}{c|}{\makecell{\textbf{BlendSQL with} \\ \texttt{gemma-4-E4B-it-FP8}}} & \multicolumn{3}{c}{\makecell{\textbf{Avg. SemBench System with} \\ \texttt{gemini-2.5-flash}}} \\
\cmidrule(lr){2-4} \cmidrule(l){5-7}
\textbf{Query} & \textbf{Lat.\,(s)} & \textbf{Quality} & \textbf{Cost\,(5\,runs)} & \textbf{Lat.\,(s)} & \textbf{Quality} & \textbf{Cost\,(5\,runs)} \\
\midrule
\href{https://github.com/CapitalOne-Research/play-by-the-type-rules/blob/main/sembench/src/queries/cars/blendsql/Q1.sql}{Q1 \textemoji} & \tikz[baseline=(X.base)]{\node[fill=bestgreen,rounded corners=3pt,inner xsep=3pt,inner ysep=1.5pt](X){\bfseries 191.33};} & 0.64 & \tikz[baseline=(X.base)]{\node[fill=bestgreen,rounded corners=3pt,inner xsep=3pt,inner ysep=1.5pt](X){\bfseries \$0.05};} & 476.75 & \tikz[baseline=(X.base)]{\node[fill=bestgreen,rounded corners=3pt,inner xsep=3pt,inner ysep=1.5pt](X){\bfseries 0.78};} & \$8.74 \\
\rowcolor{rowgray}
\href{https://github.com/CapitalOne-Research/play-by-the-type-rules/blob/main/sembench/src/queries/cars/blendsql/Q2.sql}{Q2 \audioemoji} & \tikz[baseline=(X.base)]{\node[fill=bestgreen,rounded corners=3pt,inner xsep=3pt,inner ysep=1.5pt](X){\bfseries 1.81};} & 0.00 & \tikz[baseline=(X.base)]{\node[fill=bestgreen,rounded corners=3pt,inner xsep=3pt,inner ysep=1.5pt](X){\bfseries \$5e-4};} & 10.77 & \tikz[baseline=(X.base)]{\node[fill=bestgreen,rounded corners=3pt,inner xsep=3pt,inner ysep=1.5pt](X){\bfseries 0.06};} & \$0.04 \\
\href{https://github.com/CapitalOne-Research/play-by-the-type-rules/blob/main/sembench/src/queries/cars/blendsql/Q3.sql}{Q3 \imageemoji} & \tikz[baseline=(X.base)]{\node[fill=bestgreen,rounded corners=3pt,inner xsep=3pt,inner ysep=1.5pt](X){\bfseries 1.53};} & \tikz[baseline=(X.base)]{\node[fill=bestgreen,rounded corners=3pt,inner xsep=3pt,inner ysep=1.5pt](X){\bfseries 1.00};} & \tikz[baseline=(X.base)]{\node[fill=bestgreen,rounded corners=3pt,inner xsep=3pt,inner ysep=1.5pt](X){\bfseries \$4e-4};} & 166.23 & 0.94 & \$3.78 \\
\rowcolor{rowgray}
\href{https://github.com/CapitalOne-Research/play-by-the-type-rules/blob/main/sembench/src/queries/cars/blendsql/Q4.sql}{Q4 \textemoji} & \tikz[baseline=(X.base)]{\node[fill=bestgreen,rounded corners=3pt,inner xsep=3pt,inner ysep=1.5pt](X){\bfseries 182.48};} & \tikz[baseline=(X.base)]{\node[fill=tiecolor,rounded corners=3pt,inner xsep=3pt,inner ysep=1.5pt](X){\bfseries 0.99};} & \tikz[baseline=(X.base)]{\node[fill=bestgreen,rounded corners=3pt,inner xsep=3pt,inner ysep=1.5pt](X){\bfseries \$0.05};} & 525.77 & \tikz[baseline=(X.base)]{\node[fill=tiecolor,rounded corners=3pt,inner xsep=3pt,inner ysep=1.5pt](X){\bfseries 0.99};} & \$8.59 \\
\href{https://github.com/CapitalOne-Research/play-by-the-type-rules/blob/main/sembench/src/queries/cars/blendsql/Q5.sql}{Q5 \imageemoji \audioemoji} & \tikz[baseline=(X.base)]{\node[fill=bestgreen,rounded corners=3pt,inner xsep=3pt,inner ysep=1.5pt](X){\bfseries 2.16};} & 0.00 & \tikz[baseline=(X.base)]{\node[fill=bestgreen,rounded corners=3pt,inner xsep=3pt,inner ysep=1.5pt](X){\bfseries \$5e-4};} & 182.97 & \tikz[baseline=(X.base)]{\node[fill=bestgreen,rounded corners=3pt,inner xsep=3pt,inner ysep=1.5pt](X){\bfseries 1.00};} & \$5.15 \\
\rowcolor{rowgray}
\href{https://github.com/CapitalOne-Research/play-by-the-type-rules/blob/main/sembench/src/queries/cars/blendsql/Q6.sql}{Q6 \textemoji \imageemoji \audioemoji} & \tikz[baseline=(X.base)]{\node[fill=bestgreen,rounded corners=3pt,inner xsep=3pt,inner ysep=1.5pt](X){\bfseries 215.09};} & 0.88 & \tikz[baseline=(X.base)]{\node[fill=bestgreen,rounded corners=3pt,inner xsep=3pt,inner ysep=1.5pt](X){\bfseries \$0.05};} & 415.67 & \tikz[baseline=(X.base)]{\node[fill=bestgreen,rounded corners=3pt,inner xsep=3pt,inner ysep=1.5pt](X){\bfseries 0.96};} & \$10.78 \\
\href{https://github.com/CapitalOne-Research/play-by-the-type-rules/blob/main/sembench/src/queries/cars/blendsql/Q7.sql}{Q7 \textemoji \imageemoji \audioemoji} & \tikz[baseline=(X.base)]{\node[fill=bestgreen,rounded corners=3pt,inner xsep=3pt,inner ysep=1.5pt](X){\bfseries 364.85};} & \tikz[baseline=(X.base)]{\node[fill=tiecolor,rounded corners=3pt,inner xsep=3pt,inner ysep=1.5pt](X){\bfseries 0.54};} & \tikz[baseline=(X.base)]{\node[fill=bestgreen,rounded corners=3pt,inner xsep=3pt,inner ysep=1.5pt](X){\bfseries \$0.09};} & 1038.30 & \tikz[baseline=(X.base)]{\node[fill=tiecolor,rounded corners=3pt,inner xsep=3pt,inner ysep=1.5pt](X){\bfseries 0.53};} & \$17.83 \\
\rowcolor{rowgray}
\href{https://github.com/CapitalOne-Research/play-by-the-type-rules/blob/main/sembench/src/queries/cars/blendsql/Q8.sql}{Q8 \imageemoji} & \tikz[baseline=(X.base)]{\node[fill=bestgreen,rounded corners=3pt,inner xsep=3pt,inner ysep=1.5pt](X){\bfseries 27.79};} & 0.21 & \tikz[baseline=(X.base)]{\node[fill=bestgreen,rounded corners=3pt,inner xsep=3pt,inner ysep=1.5pt](X){\bfseries \$7e-3};} & 431.18 & \tikz[baseline=(X.base)]{\node[fill=bestgreen,rounded corners=3pt,inner xsep=3pt,inner ysep=1.5pt](X){\bfseries 0.29};} & \$7.12 \\
\href{https://github.com/CapitalOne-Research/play-by-the-type-rules/blob/main/sembench/src/queries/cars/blendsql/Q10.sql}{Q10 \textemoji} & \tikz[baseline=(X.base)]{\node[fill=bestgreen,rounded corners=3pt,inner xsep=3pt,inner ysep=1.5pt](X){\bfseries 351.99};} & \tikz[baseline=(X.base)]{\node[fill=tiecolor,rounded corners=3pt,inner xsep=3pt,inner ysep=1.5pt](X){\bfseries 0.50};} & \tikz[baseline=(X.base)]{\node[fill=bestgreen,rounded corners=3pt,inner xsep=3pt,inner ysep=1.5pt](X){\bfseries \$0.09};} & 425.00 & \tikz[baseline=(X.base)]{\node[fill=tiecolor,rounded corners=3pt,inner xsep=3pt,inner ysep=1.5pt](X){\bfseries 0.50};} & \$17.47 \\
\midrule
\textbf{Avg.} & \tikz[baseline=(X.base)]{\node[fill=bestgreen,rounded corners=3pt,inner xsep=3pt,inner ysep=1.5pt](X){\bfseries 148.78};} & 0.53 & \tikz[baseline=(X.base)]{\node[fill=bestgreen,rounded corners=3pt,inner xsep=3pt,inner ysep=1.5pt](X){\bfseries \$0.04};} & 408.07 & \tikz[baseline=(X.base)]{\node[fill=bestgreen,rounded corners=3pt,inner xsep=3pt,inner ysep=1.5pt](X){\bfseries 0.67};} & \$8.83 \\
\textbf{Avg.\ Std.} & 0.54 & 0.00 & \$1e-4 & - & - & \- \\
\textbf{Wins} & 9 & 1 & 9 & 0 & 5 & 0 \\
\textbf{Total Cost} & & & \tikz[baseline=(X.base)]{\node[fill=bestgreen,rounded corners=3pt,inner xsep=3pt,inner ysep=1.5pt](X){\bfseries \$0.33};} & & & \$79.51 \\
\bottomrule
\end{tabular}

%% file: tables/wildlife.tex
\footnotesize
\label{tab:blendsql_vs_baseline_gemma_e4b}
\begin{tabular}{l rrr | rrr}
\toprule
 & \multicolumn{3}{c|}{\makecell{\textbf{BlendSQL with} \\ \texttt{gemma-4-E4B-it-FP8}}} & \multicolumn{3}{c}{\makecell{\textbf{Avg. SemBench System with} \\ \texttt{gemini-2.5-flash}}} \\
\cmidrule(lr){2-4} \cmidrule(l){5-7}
\textbf{Query} & \textbf{Lat.\,(s)} & \textbf{Quality} & \textbf{Cost\,(5\,runs)} & \textbf{Lat.\,(s)} & \textbf{Quality} & \textbf{Cost\,(5\,runs)} \\
\midrule
\href{https://github.com/CapitalOne-Research/play-by-the-type-rules/blob/main/sembench/src/queries/wildlife/blendsql/Q1.sql}{Q1 \imageemoji} & \tikz[baseline=(X.base)]{\node[fill=bestgreen,rounded corners=3pt,inner xsep=3pt,inner ysep=1.5pt](X){\bfseries 9.41};} & \tikz[baseline=(X.base)]{\node[fill=tiecolor,rounded corners=3pt,inner xsep=3pt,inner ysep=1.5pt](X){\bfseries 0.78};} & \tikz[baseline=(X.base)]{\node[fill=bestgreen,rounded corners=3pt,inner xsep=3pt,inner ysep=1.5pt](X){\bfseries \$2e-3};} & 44.20 & \tikz[baseline=(X.base)]{\node[fill=tiecolor,rounded corners=3pt,inner xsep=3pt,inner ysep=1.5pt](X){\bfseries 0.79};} & \$0.57 \\
\rowcolor{rowgray}
\href{https://github.com/CapitalOne-Research/play-by-the-type-rules/blob/main/sembench/src/queries/wildlife/blendsql/Q2.sql}{Q2 \audioemoji} & \tikz[baseline=(X.base)]{\node[fill=bestgreen,rounded corners=3pt,inner xsep=3pt,inner ysep=1.5pt](X){\bfseries 4.16};} & 0.00 & \tikz[baseline=(X.base)]{\node[fill=bestgreen,rounded corners=3pt,inner xsep=3pt,inner ysep=1.5pt](X){\bfseries \$1e-3};} & 5.57 & \tikz[baseline=(X.base)]{\node[fill=bestgreen,rounded corners=3pt,inner xsep=3pt,inner ysep=1.5pt](X){\bfseries 0.17};} & \$0.05 \\
\href{https://github.com/CapitalOne-Research/play-by-the-type-rules/blob/main/sembench/src/queries/wildlife/blendsql/Q3.sql}{Q3 \imageemoji} & \tikz[baseline=(X.base)]{\node[fill=bestgreen,rounded corners=3pt,inner xsep=3pt,inner ysep=1.5pt](X){\bfseries 9.29};} & 0.20 & \tikz[baseline=(X.base)]{\node[fill=bestgreen,rounded corners=3pt,inner xsep=3pt,inner ysep=1.5pt](X){\bfseries \$2e-3};} & 38.02 & \tikz[baseline=(X.base)]{\node[fill=bestgreen,rounded corners=3pt,inner xsep=3pt,inner ysep=1.5pt](X){\bfseries 0.25};} & \$0.47 \\
\rowcolor{rowgray}
\href{https://github.com/CapitalOne-Research/play-by-the-type-rules/blob/main/sembench/src/queries/wildlife/blendsql/Q4.sql}{Q4 \audioemoji} & \tikz[baseline=(X.base)]{\node[fill=bestgreen,rounded corners=3pt,inner xsep=3pt,inner ysep=1.5pt](X){\bfseries 4.01};} & \tikz[baseline=(X.base)]{\node[fill=bestgreen,rounded corners=3pt,inner xsep=3pt,inner ysep=1.5pt](X){\bfseries 0.40};} & \tikz[baseline=(X.base)]{\node[fill=bestgreen,rounded corners=3pt,inner xsep=3pt,inner ysep=1.5pt](X){\bfseries \$1e-3};} & 4.47 & 0.33 & \$0.04 \\
\href{https://github.com/CapitalOne-Research/play-by-the-type-rules/blob/main/sembench/src/queries/wildlife/blendsql/Q5.sql}{Q5 \imageemoji \audioemoji} & 13.39 & \tikz[baseline=(X.base)]{\node[fill=bestgreen,rounded corners=3pt,inner xsep=3pt,inner ysep=1.5pt](X){\bfseries 0.93};} & \tikz[baseline=(X.base)]{\node[fill=bestgreen,rounded corners=3pt,inner xsep=3pt,inner ysep=1.5pt](X){\bfseries \$3e-3};} & \tikz[baseline=(X.base)]{\node[fill=bestgreen,rounded corners=3pt,inner xsep=3pt,inner ysep=1.5pt](X){\bfseries 11.67};} & 0.75 & \$0.43 \\
\rowcolor{rowgray}
\href{https://github.com/CapitalOne-Research/play-by-the-type-rules/blob/main/sembench/src/queries/wildlife/blendsql/Q6.sql}{Q6 \imageemoji \audioemoji} & \tikz[baseline=(X.base)]{\node[fill=bestgreen,rounded corners=3pt,inner xsep=3pt,inner ysep=1.5pt](X){\bfseries 13.34};} & 0.20 & \tikz[baseline=(X.base)]{\node[fill=bestgreen,rounded corners=3pt,inner xsep=3pt,inner ysep=1.5pt](X){\bfseries \$3e-3};} & 18.93 & \tikz[baseline=(X.base)]{\node[fill=bestgreen,rounded corners=3pt,inner xsep=3pt,inner ysep=1.5pt](X){\bfseries 0.23};} & \$0.52 \\
\href{https://github.com/CapitalOne-Research/play-by-the-type-rules/blob/main/sembench/src/queries/wildlife/blendsql/Q7.sql}{Q7 \imageemoji} & \tikz[baseline=(X.base)]{\node[fill=bestgreen,rounded corners=3pt,inner xsep=3pt,inner ysep=1.5pt](X){\bfseries 18.58};} & \tikz[baseline=(X.base)]{\node[fill=tiecolor,rounded corners=3pt,inner xsep=3pt,inner ysep=1.5pt](X){\bfseries 1.00};} & \tikz[baseline=(X.base)]{\node[fill=bestgreen,rounded corners=3pt,inner xsep=3pt,inner ysep=1.5pt](X){\bfseries \$5e-3};} & 71.40 & \tikz[baseline=(X.base)]{\node[fill=tiecolor,rounded corners=3pt,inner xsep=3pt,inner ysep=1.5pt](X){\bfseries 1.00};} & \$0.98 \\
\rowcolor{rowgray}
\href{https://github.com/CapitalOne-Research/play-by-the-type-rules/blob/main/sembench/src/queries/wildlife/blendsql/Q8.sql}{Q8 \imageemoji \audioemoji} & \tikz[baseline=(X.base)]{\node[fill=bestgreen,rounded corners=3pt,inner xsep=3pt,inner ysep=1.5pt](X){\bfseries 26.68};} & \tikz[baseline=(X.base)]{\node[fill=bestgreen,rounded corners=3pt,inner xsep=3pt,inner ysep=1.5pt](X){\bfseries 0.83};} & \tikz[baseline=(X.base)]{\node[fill=bestgreen,rounded corners=3pt,inner xsep=3pt,inner ysep=1.5pt](X){\bfseries \$7e-3};} & 31.40 & 0.75 & \$0.80 \\
\href{https://github.com/CapitalOne-Research/play-by-the-type-rules/blob/main/sembench/src/queries/wildlife/blendsql/Q9.sql}{Q9 \imageemoji \audioemoji} & \tikz[baseline=(X.base)]{\node[fill=bestgreen,rounded corners=3pt,inner xsep=3pt,inner ysep=1.5pt](X){\bfseries 11.92};} & \tikz[baseline=(X.base)]{\node[fill=tiecolor,rounded corners=3pt,inner xsep=3pt,inner ysep=1.5pt](X){\bfseries 0.61};} & \tikz[baseline=(X.base)]{\node[fill=bestgreen,rounded corners=3pt,inner xsep=3pt,inner ysep=1.5pt](X){\bfseries \$3e-3};} & 24.63 & \tikz[baseline=(X.base)]{\node[fill=tiecolor,rounded corners=3pt,inner xsep=3pt,inner ysep=1.5pt](X){\bfseries 0.61};} & \$0.55 \\
\rowcolor{rowgray}
\href{https://github.com/CapitalOne-Research/play-by-the-type-rules/blob/main/sembench/src/queries/wildlife/blendsql/Q10.sql}{Q10 \imageemoji} & \tikz[baseline=(X.base)]{\node[fill=bestgreen,rounded corners=3pt,inner xsep=3pt,inner ysep=1.5pt](X){\bfseries 9.27};} & \tikz[baseline=(X.base)]{\node[fill=bestgreen,rounded corners=3pt,inner xsep=3pt,inner ysep=1.5pt](X){\bfseries 0.40};} & \tikz[baseline=(X.base)]{\node[fill=bestgreen,rounded corners=3pt,inner xsep=3pt,inner ysep=1.5pt](X){\bfseries \$2e-3};} & 38.02 & 0.25 & \$0.47 \\
\midrule
\textbf{Avg.} & \tikz[baseline=(X.base)]{\node[fill=bestgreen,rounded corners=3pt,inner xsep=3pt,inner ysep=1.5pt](X){\bfseries 12.01};} & \tikz[baseline=(X.base)]{\node[fill=bestgreen,rounded corners=3pt,inner xsep=3pt,inner ysep=1.5pt](X){\bfseries 0.54};} & \tikz[baseline=(X.base)]{\node[fill=bestgreen,rounded corners=3pt,inner xsep=3pt,inner ysep=1.5pt](X){\bfseries \$3e-3};} & 28.83 & 0.51 & \$0.49 \\
\textbf{Avg.\ Std.} & 0.08 & 0.22 & \$2e-5 & - & - & - \\
\textbf{Wins} & 9 & 4 & 10 & 1 & 3 & 0 \\
\textbf{Total Cost} & & & \tikz[baseline=(X.base)]{\node[fill=bestgreen,rounded corners=3pt,inner xsep=3pt,inner ysep=1.5pt](X){\bfseries \$0.03};} & & & \$4.88 \\
\bottomrule
\end{tabular}

%% file: sections/5_related_work.tex
\section{Related Work}

\paragraph{Combining Language Models with Database Systems}

Combining language models with structured data operators is a widely studied topic. The origin of such an approach can be traced back to early SQL operators enabling statistical machine learning and data mining \cite{meo1996new, han1996dmql, imielinski1999msql}. Later, database management systems with native predictive functions utilizing models such as logistic regression and support vector machines became more common, such as SQL Server's \inlinesql{PREDICT} function in 2017 and BigQuery ML in 2018 \cite{hellerstein2012madlib, googlebigquerypredict, sqlserver2017}. To the best of our knowledge, \citet{bae2023ehrxqa} were the first to propose the idea of putting calls to a multi-modal neural model into a SQL query. Others have since continued exploration into domain-specific languages for combining the generalized computations of language models with the structured reasoning of traditional database query languages~\citep{cheng2023binding,dorbanibeyond,tjangnakasuql,patel2024semanticoperators,glennblendsql}.

Modern approaches integrate language models with database management systems at varying levels. While \citet{patel2024semanticoperators} intervenes via a Pandas API, \citet{dorbanibeyond} build out a set of custom UDFs for the online analytical processing DBMS DuckDB \citep{raasveldt2019duckdb}. \citet{tjangnakasuql} build out UDFs for the PostgreSQL DBMS \citep{postgresql}, with additional calls to language models to determine the semantic equivalency LM-generated values against native database values. Palimpzest \cite{liu2025palimpzest} is unique in its formulation of the cost optimization problem, allowing the system to search over a space of various-priced closed LM APIs. A subset of work specifically explores efficient methods for optimizing LM functions in relational systems \citep{kim2024optimizing, liu2024optimizing}.

\paragraph{Constrained Decoding}

Constrained decoding refers to the process of controlling the output of language models by applying masks at the decoding level, such that generations adhere to a specific pre-determined constraint \citep{deutsch2019general}. These constraints are typically encoded via regular expressions or context-free grammars, and optimized decoding engines have emerged for deriving and applying masks \citep{willard2023efficient,geng2023grammar,park2025flexible,dong2024xgrammar,guidance}. Recent work has explored alternatives to the predominant locally-constrained decoding paradigm \cite{lipkin2025fast}. Most relevant to our work is \citet{mundler2025type}, who present an algorithm to enforce the well-typedness of LLM-generated TypeScript code. Whereas they tackle the problem of determining whether a partial program can be completed into a well-typed program, we explore constraints for integrating LM outputs into a declarative query language like SQL. 

%% file: sections/6_conclusion.tex
\section{Conclusion}
Under token-based API pricing, cost scales with the number of tokens processed, independent of inference speed. Open-weight models on local, hourly-billed hardware invert this relationship: by tuning parameters such as max concurrency (Figure \ref{fig:concurrency_plot}) and query optimizations (Figure \ref{fig:feature_ablation_plot}), faster inference directly translates to lower cost. This shift in incentive structure allows researchers to focus on core inference innovations with known architectures and hardware, rather than engineering around black-box models, arbitrary rate limits, and opaque infrastructure. The cost difference becomes stark at scale and massively improves the accessibility of LM-DB research: \citet{lao2025sembench} report spending an average of \$2,988.32 evaluating \textsc{Palimpzest} on the \textsc{ecomm} scenario five times across five scale factors. We show that the equivalent workload with \textsc{BlendSQL} and Gemma 4 E4B costs \$6.96. Perhaps most importantly, numerous studies have shown API quality fluctuates over time, making full reproducibility impossible \cite{chen2024chatgpt,tu2023chatlog,angermeir2025reflections}. Open-weight models guarantee full reproducibility by decoupling experimental results from the volatility of proprietary LM endpoints. 

We presented \textsc{BlendSQL} v0.1.0, a LM-DB system that elevates small, open-weight language models to performance competitive with closed API solutions. We demonstrated a cost reduction of 390x and latency reduction of 3.8x, with performance matching or exceeding systems with closed-weight models on text and image inputs. We highlighted the modality gap between small open-weight and closed-weight models, demonstrating a stark quality difference on audio data. Finally, we examined the components aiding the success of LM-DB systems through a series of ablations. 
% Additionally, it is difficult to draw meaningful conclusions from comparative experiments. Our study finds that small open-weight models can match or outperform closed-source counterparts on tasks involving text and image modalities, but fall behind on audio-based tasks. Lack of details on model Why might this be the case? Pricing docs show that \texttt{gemini-2.5-flash} operates on a cost model of \$1.00 per million audio tokens, and \$0.30 per million tokens of all other modalities \cite{geminipricing}. The reason for this is unclear: Is this due to specialized encoder stages? Do audio requests get routed to a specialized model? Closed source APIs make this impossible to tell. \cite{comanici2025gemini}

%% file: sections/7_appendix.tex
\section{Appendix}

\subsection{vLLM Config}

All local vLLM servers were launched using the following command with \texttt{vLLM==0.21.0}

\begin{lstlisting}[language=bash]
vllm serve "${model_path}" \
--host 0.0.0.0 \
--port 8000 \
--tensor-parallel-size 1 \
--enable-prefix-caching \
--max-model-len 32000 \
--structured-outputs-config.backend guidance \
--gpu_memory_utilization 0.9 \
--mm-processor-kwargs '{"max_length": 480000}' \
--enable-prompt-tokens-details
\end{lstlisting}

\subsection{Prompts}

\begin{tcolorbox}[fonttitle=\fontfamily{pbk}\selectfont\bfseries,
fontupper=\fontsize{8}{9}\fontfamily{ppl}\selectfont,
fontlower=\fontfamily{cmtt}\selectfont\scshape,
title=\textsc{llmmap} Prompt,
width=\linewidth,
arc=1mm, auto outer arc]
\begin{Verbatim}[commandchars=\\\{\}]
You are a helpful assistant. You will be 
presented with some context and a question. 
\blue{\{\{return\_type\_instructions\}\}} 

\blue{\{% if quantifier is not none %\}}
\blue{\{\{quantifier\_disclaimer\}\}}
\blue{\{% endif % \}}

An example is shown below.

\blue{\{\{one\_shot\_example\}\}}

---

QUESTION:
\blue{\{\{question\}\}}

CONTEXT:
\blue{\{\{context\_as\_dict\}\}}

\blue{\{% if options is not none %\}}
OPTIONS:
\blue{\{\{options\}\}}
\blue{\{% endif % \}}

ANSWER:
\end{Verbatim}
\end{tcolorbox}

\begin{tcolorbox}[fonttitle=\fontfamily{pbk}\selectfont\bfseries,
fontupper=\fontsize{8}{9}\fontfamily{ppl}\selectfont,
fontlower=\fontfamily{cmtt}\selectfont\scshape,
title=\textsc{llmqa} Prompt,
width=\linewidth,
arc=1mm, auto outer arc]
\begin{Verbatim}[commandchars=\\\{\}]
Answer the question given the context, if provided.
Keep the answers as short as possible, 
without leading context. For example, do not 
say 'The answer is 2', simply say '2'. 

Your response format should match the specified 
'Return type', if provided. 

Question: \blue{\{\{question\}\}}

\blue{\{% if context is not none %\}}
Context: \blue{\{\{context\_as\_dict\}\}}
\blue{\{% endif % \}}

\blue{\{% if options is not none %\}}
Options: \blue{\{\{options\}\}}
\blue{\{% endif % \}}

\blue{\{% if return\_type is not none %\}}
Return type: \blue{\{\{return\_type\}\}}
\blue{\{% endif % \}}

Answer:
\end{Verbatim}
\end{tcolorbox}

\begin{figure*}[t]
\caption{Full SemBench results. \textsc{BlendSQL} uses \texttt{gemma-4-E4B-FP8}, other systems use \texttt{gemini-2.5-flash}.}
\label{tab:full_sembench_results}
\begin{minipage}[t]{\textwidth}
  \centering
  \subcaption{Full \textsc{mmqa} results.}
  \resizebox{\textwidth}{!}{\input{tables/full_mmqa}}
\end{minipage}

\vspace{1em}

\begin{minipage}[t]{\textwidth}
  \subcaption{Full \textsc{ecomm} results.}
  \centering
  \resizebox{\textwidth}{!}{\input{tables/full_ecomm}}
\end{minipage}

\vspace{1em}

\begin{minipage}[t]{\textwidth}
  \subcaption{Full \textsc{movie} results.}
  \centering
  \resizebox{\textwidth}{!}{\input{tables/full_movie}}
\end{minipage}

\end{figure*}

\begin{figure*}[t]

\begin{minipage}[t]{\textwidth}
  \subcaption{Full \textsc{cars} results.}
  \centering
  \resizebox{\textwidth}{!}{\input{tables/full_cars}}
\end{minipage}

\vspace{1em}

\begin{minipage}[t]{\textwidth}
  \centering
  \subcaption{Full \textsc{wildlife} results.}
  \resizebox{\textwidth}{!}{\input{tables/full_wildlife}}
\end{minipage}

\end{figure*}

\begin{figure*}[h]
    \centering
     \includegraphics[scale=0.6]{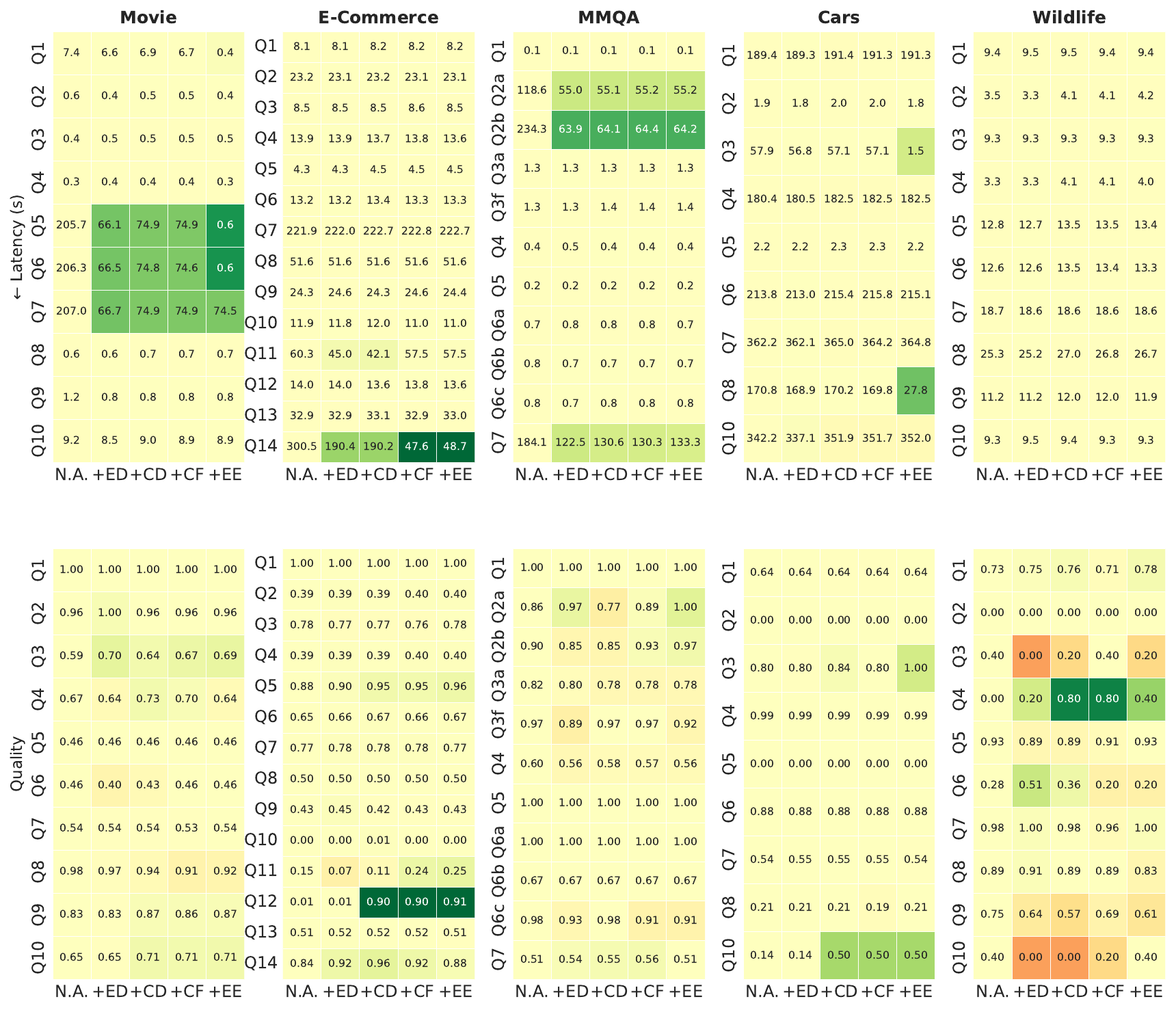}
    \caption{Plotting the query-level latency and quality changes on SemBench scenarios with \textsc{BlendSQL} and Gemma 4 E4B under various feature ablations.}
    \label{fig:ablation_heatmap}
\end{figure*}

%% file: tables/full_mmqa.tex
\footnotesize
\label{tab:blendsql_vs_baseline_gemma_e4b}
\begin{tabular}{l rrr | rrr | rrr | rrr | rrr}
\toprule
  & \multicolumn{3}{c|}{\textbf{BlendSQL}} & \multicolumn{3}{c|}{\textbf{BigQuery}} & \multicolumn{3}{c|}{\textbf{LOTUS}} & \multicolumn{3}{c|}{\textbf{Palimpzest}} & \multicolumn{3}{c}{\textbf{ThalamusDB}} \\
\cmidrule(lr){2-4} \cmidrule(lr){5-7} \cmidrule(lr){8-10} \cmidrule(lr){11-13} \cmidrule(l){14-16}
\textbf{Query} & \textbf{Lat.\,(s)} & \textbf{Quality} & \textbf{Cost\,(5\,runs)} & \textbf{Lat.\,(s)} & \textbf{Quality} & \textbf{Cost\,(5\,runs)} & \textbf{Lat.\,(s)} & \textbf{Quality} & \textbf{Cost\,(5\,runs)} & \textbf{Lat.\,(s)} & \textbf{Quality} & \textbf{Cost\,(5\,runs)} & \textbf{Lat.\,(s)} & \textbf{Quality} & \textbf{Cost\,(5\,runs)} \\
\midrule
Q1 & \tikz[baseline=(X.base)]{\node[fill=bestgreen,rounded corners=3pt,inner xsep=3pt,inner ysep=1.5pt](X){\bfseries 0.07};} & \tikz[baseline=(X.base)]{\node[fill=tiecolor,rounded corners=3pt,inner xsep=3pt,inner ysep=1.5pt](X){\bfseries 1.00};} & \tikz[baseline=(X.base)]{\node[fill=bestgreen,rounded corners=3pt,inner xsep=3pt,inner ysep=1.5pt](X){\bfseries \$2e-5};} & 14.10 & \tikz[baseline=(X.base)]{\node[fill=tiecolor,rounded corners=3pt,inner xsep=3pt,inner ysep=1.5pt](X){\bfseries 1.00};} & \$0.05 & 7.40 & \tikz[baseline=(X.base)]{\node[fill=tiecolor,rounded corners=3pt,inner xsep=3pt,inner ysep=1.5pt](X){\bfseries 1.00};} & \$0.10 & 4.50 & \tikz[baseline=(X.base)]{\node[fill=tiecolor,rounded corners=3pt,inner xsep=3pt,inner ysep=1.5pt](X){\bfseries 1.00};} & \$0.01 & -- & -- & -- \\
\rowcolor{rowgray}
Q2a & 55.15 & \tikz[baseline=(X.base)]{\node[fill=tiecolor,rounded corners=3pt,inner xsep=3pt,inner ysep=1.5pt](X){\bfseries 1.00};} & \tikz[baseline=(X.base)]{\node[fill=bestgreen,rounded corners=3pt,inner xsep=3pt,inner ysep=1.5pt](X){\bfseries \$0.01};} & \tikz[baseline=(X.base)]{\node[fill=bestgreen,rounded corners=3pt,inner xsep=3pt,inner ysep=1.5pt](X){\bfseries 53.80};} & 0.00 & \$0.40 & 169.60 & 0.83 & \$4.45 & 135.90 & \tikz[baseline=(X.base)]{\node[fill=tiecolor,rounded corners=3pt,inner xsep=3pt,inner ysep=1.5pt](X){\bfseries 1.00};} & \$5.20 & -- & -- & -- \\
Q2b & 64.22 & 0.97 & \tikz[baseline=(X.base)]{\node[fill=bestgreen,rounded corners=3pt,inner xsep=3pt,inner ysep=1.5pt](X){\bfseries \$0.02};} & \tikz[baseline=(X.base)]{\node[fill=bestgreen,rounded corners=3pt,inner xsep=3pt,inner ysep=1.5pt](X){\bfseries 38.40};} & 0.00 & \$0.60 & 166.80 & 0.83 & \$4.45 & 133.80 & \tikz[baseline=(X.base)]{\node[fill=bestgreen,rounded corners=3pt,inner xsep=3pt,inner ysep=1.5pt](X){\bfseries 1.00};} & \$5.20 & -- & -- & -- \\
\rowcolor{rowgray}
Q3a & \tikz[baseline=(X.base)]{\node[fill=bestgreen,rounded corners=3pt,inner xsep=3pt,inner ysep=1.5pt](X){\bfseries 1.35};} & 0.78 & \tikz[baseline=(X.base)]{\node[fill=bestgreen,rounded corners=3pt,inner xsep=3pt,inner ysep=1.5pt](X){\bfseries \$3e-4};} & 19.60 & 0.72 & \$0.05 & 4.20 & 0.83 & \$0.05 & 4.50 & \tikz[baseline=(X.base)]{\node[fill=bestgreen,rounded corners=3pt,inner xsep=3pt,inner ysep=1.5pt](X){\bfseries 1.00};} & \$5.20 & 11.00 & 0.75 & \$0.05 \\
Q3f & \tikz[baseline=(X.base)]{\node[fill=bestgreen,rounded corners=3pt,inner xsep=3pt,inner ysep=1.5pt](X){\bfseries 1.41};} & 0.92 & \tikz[baseline=(X.base)]{\node[fill=bestgreen,rounded corners=3pt,inner xsep=3pt,inner ysep=1.5pt](X){\bfseries \$4e-4};} & 22.30 & 0.67 & \$0.05 & 4.20 & \tikz[baseline=(X.base)]{\node[fill=tiecolor,rounded corners=3pt,inner xsep=3pt,inner ysep=1.5pt](X){\bfseries 1.00};} & \$0.05 & 8.00 & \tikz[baseline=(X.base)]{\node[fill=tiecolor,rounded corners=3pt,inner xsep=3pt,inner ysep=1.5pt](X){\bfseries 1.00};} & \$0.10 & 7.00 & \tikz[baseline=(X.base)]{\node[fill=tiecolor,rounded corners=3pt,inner xsep=3pt,inner ysep=1.5pt](X){\bfseries 1.00};} & \$0.05 \\
\rowcolor{rowgray}
Q4 & \tikz[baseline=(X.base)]{\node[fill=bestgreen,rounded corners=3pt,inner xsep=3pt,inner ysep=1.5pt](X){\bfseries 0.45};} & 0.56 & \tikz[baseline=(X.base)]{\node[fill=bestgreen,rounded corners=3pt,inner xsep=3pt,inner ysep=1.5pt](X){\bfseries \$1e-4};} & 9.70 & \tikz[baseline=(X.base)]{\node[fill=bestgreen,rounded corners=3pt,inner xsep=3pt,inner ysep=1.5pt](X){\bfseries 0.60};} & \$0.01 & -- & -- & -- & 1.20 & 0.54 & \$0.03 & -- & -- & -- \\
Q5 & \tikz[baseline=(X.base)]{\node[fill=bestgreen,rounded corners=3pt,inner xsep=3pt,inner ysep=1.5pt](X){\bfseries 0.24};} & \tikz[baseline=(X.base)]{\node[fill=tiecolor,rounded corners=3pt,inner xsep=3pt,inner ysep=1.5pt](X){\bfseries 1.00};} & \tikz[baseline=(X.base)]{\node[fill=tiecolor,rounded corners=3pt,inner xsep=3pt,inner ysep=1.5pt](X){\bfseries \$6e-5};} & -- & -- & -- & 0.48 & \tikz[baseline=(X.base)]{\node[fill=tiecolor,rounded corners=3pt,inner xsep=3pt,inner ysep=1.5pt](X){\bfseries 1.00};} & \tikz[baseline=(X.base)]{\node[fill=tiecolor,rounded corners=3pt,inner xsep=3pt,inner ysep=1.5pt](X){\bfseries \$5e-3};} & 0.50 & \tikz[baseline=(X.base)]{\node[fill=tiecolor,rounded corners=3pt,inner xsep=3pt,inner ysep=1.5pt](X){\bfseries 1.00};} & \tikz[baseline=(X.base)]{\node[fill=tiecolor,rounded corners=3pt,inner xsep=3pt,inner ysep=1.5pt](X){\bfseries \$5e-3};} & -- & -- & -- \\
\rowcolor{rowgray}
Q6a & \tikz[baseline=(X.base)]{\node[fill=bestgreen,rounded corners=3pt,inner xsep=3pt,inner ysep=1.5pt](X){\bfseries 0.73};} & \tikz[baseline=(X.base)]{\node[fill=tiecolor,rounded corners=3pt,inner xsep=3pt,inner ysep=1.5pt](X){\bfseries 1.00};} & \tikz[baseline=(X.base)]{\node[fill=bestgreen,rounded corners=3pt,inner xsep=3pt,inner ysep=1.5pt](X){\bfseries \$2e-4};} & 18.90 & 0.03 & \$0.02 & 4.40 & \tikz[baseline=(X.base)]{\node[fill=tiecolor,rounded corners=3pt,inner xsep=3pt,inner ysep=1.5pt](X){\bfseries 1.00};} & \$0.05 & 4.00 & \tikz[baseline=(X.base)]{\node[fill=tiecolor,rounded corners=3pt,inner xsep=3pt,inner ysep=1.5pt](X){\bfseries 1.00};} & \$0.10 & 5.60 & 0.33 & \$0.01 \\
Q6b & \tikz[baseline=(X.base)]{\node[fill=bestgreen,rounded corners=3pt,inner xsep=3pt,inner ysep=1.5pt](X){\bfseries 0.71};} & 0.67 & \tikz[baseline=(X.base)]{\node[fill=bestgreen,rounded corners=3pt,inner xsep=3pt,inner ysep=1.5pt](X){\bfseries \$2e-4};} & 17.30 & 0.04 & \$0.02 & 3.80 & \tikz[baseline=(X.base)]{\node[fill=tiecolor,rounded corners=3pt,inner xsep=3pt,inner ysep=1.5pt](X){\bfseries 1.00};} & \$0.05 & 4.10 & \tikz[baseline=(X.base)]{\node[fill=tiecolor,rounded corners=3pt,inner xsep=3pt,inner ysep=1.5pt](X){\bfseries 1.00};} & \$0.10 & 5.50 & \tikz[baseline=(X.base)]{\node[fill=tiecolor,rounded corners=3pt,inner xsep=3pt,inner ysep=1.5pt](X){\bfseries 1.00};} & \$0.01 \\
\rowcolor{rowgray}
Q6c & \tikz[baseline=(X.base)]{\node[fill=bestgreen,rounded corners=3pt,inner xsep=3pt,inner ysep=1.5pt](X){\bfseries 0.83};} & 0.91 & \tikz[baseline=(X.base)]{\node[fill=bestgreen,rounded corners=3pt,inner xsep=3pt,inner ysep=1.5pt](X){\bfseries \$2e-4};} & 17.40 & 0.13 & \$0.02 & 4.60 & \tikz[baseline=(X.base)]{\node[fill=tiecolor,rounded corners=3pt,inner xsep=3pt,inner ysep=1.5pt](X){\bfseries 1.00};} & \$0.05 & 4.70 & \tikz[baseline=(X.base)]{\node[fill=tiecolor,rounded corners=3pt,inner xsep=3pt,inner ysep=1.5pt](X){\bfseries 1.00};} & \$0.10 & 13.40 & 0.53 & \$0.01 \\
Q7 & 133.26 & \tikz[baseline=(X.base)]{\node[fill=bestgreen,rounded corners=3pt,inner xsep=3pt,inner ysep=1.5pt](X){\bfseries 0.51};} & \tikz[baseline=(X.base)]{\node[fill=bestgreen,rounded corners=3pt,inner xsep=3pt,inner ysep=1.5pt](X){\bfseries \$0.03};} & \tikz[baseline=(X.base)]{\node[fill=bestgreen,rounded corners=3pt,inner xsep=3pt,inner ysep=1.5pt](X){\bfseries 91.70};} & 0.00 & \$5.90 & 2311.40 & 0.32 & \$68.05 & 2101.60 & 0.31 & \$78.25 & -- & -- & -- \\
\midrule
\textbf{Avg.} & 23.49 & 0.85 & \tikz[baseline=(X.base)]{\node[fill=bestgreen,rounded corners=3pt,inner xsep=3pt,inner ysep=1.5pt](X){\bfseries \$6e-3};} & 32.61 & 0.29 & \$0.79 & 267.27 & 0.88 & \$7.73 & 218.44 & \tikz[baseline=(X.base)]{\node[fill=bestgreen,rounded corners=3pt,inner xsep=3pt,inner ysep=1.5pt](X){\bfseries 0.90};} & \$8.57 & \tikz[baseline=(X.base)]{\node[fill=bestgreen,rounded corners=3pt,inner xsep=3pt,inner ysep=1.5pt](X){\bfseries 8.50};} & 0.72 & \$0.03 \\
\textbf{Avg.\ Std.} & 0.11 & 0.03 & \$3e-5 & -- & -- & -- & -- & -- & -- & -- & -- & -- & -- & -- & -- \\
\textbf{Wins} & 8 & 1 & 10 & 3 & 1 & 0 & 0 & 0 & 0 & 0 & 2 & 0 & 0 & 0 & 0 \\
\textbf{Total Cost} & & & \tikz[baseline=(X.base)]{\node[fill=bestgreen,rounded corners=3pt,inner xsep=3pt,inner ysep=1.5pt](X){\bfseries \$0.06};} & & & \$7.11 & & & \$77.30 & & & \$94.30 & & & \$0.13 \\
\bottomrule
\end{tabular}

%% file: tables/full_ecomm.tex
\footnotesize
\label{tab:blendsql_vs_baseline_gemma_e4b}
\begin{tabular}{l rrr | rrr | rrr | rrr | rrr}
\toprule
  & \multicolumn{3}{c|}{\textbf{BlendSQL}} & \multicolumn{3}{c|}{\textbf{BigQuery}} & \multicolumn{3}{c|}{\textbf{LOTUS}} & \multicolumn{3}{c|}{\textbf{Palimpzest}} & \multicolumn{3}{c}{\textbf{ThalamusDB}} \\
\cmidrule(lr){2-4} \cmidrule(lr){5-7} \cmidrule(lr){8-10} \cmidrule(lr){11-13} \cmidrule(l){14-16}
\textbf{Query} & \textbf{Lat.\,(s)} & \textbf{Quality} & \textbf{Cost\,(5\,runs)} & \textbf{Lat.\,(s)} & \textbf{Quality} & \textbf{Cost\,(5\,runs)} & \textbf{Lat.\,(s)} & \textbf{Quality} & \textbf{Cost\,(5\,runs)} & \textbf{Lat.\,(s)} & \textbf{Quality} & \textbf{Cost\,(5\,runs)} & \textbf{Lat.\,(s)} & \textbf{Quality} & \textbf{Cost\,(5\,runs)} \\
\midrule
Q1 & \tikz[baseline=(X.base)]{\node[fill=bestgreen,rounded corners=3pt,inner xsep=3pt,inner ysep=1.5pt](X){\bfseries 8.18};} & \tikz[baseline=(X.base)]{\node[fill=tiecolor,rounded corners=3pt,inner xsep=3pt,inner ysep=1.5pt](X){\bfseries 1.00};} & \tikz[baseline=(X.base)]{\node[fill=bestgreen,rounded corners=3pt,inner xsep=3pt,inner ysep=1.5pt](X){\bfseries \$2e-3};} & 21.20 & 0.59 & \$0.20 & 12.20 & \tikz[baseline=(X.base)]{\node[fill=tiecolor,rounded corners=3pt,inner xsep=3pt,inner ysep=1.5pt](X){\bfseries 1.00};} & \$0.30 & 12.20 & \tikz[baseline=(X.base)]{\node[fill=tiecolor,rounded corners=3pt,inner xsep=3pt,inner ysep=1.5pt](X){\bfseries 1.00};} & \$0.40 & 34.80 & \tikz[baseline=(X.base)]{\node[fill=tiecolor,rounded corners=3pt,inner xsep=3pt,inner ysep=1.5pt](X){\bfseries 1.00};} & \$0.15 \\
\rowcolor{rowgray}
Q2 & \tikz[baseline=(X.base)]{\node[fill=bestgreen,rounded corners=3pt,inner xsep=3pt,inner ysep=1.5pt](X){\bfseries 23.15};} & 0.40 & \tikz[baseline=(X.base)]{\node[fill=bestgreen,rounded corners=3pt,inner xsep=3pt,inner ysep=1.5pt](X){\bfseries \$6e-3};} & 55.70 & 0.21 & \$19.80 & 166.10 & \tikz[baseline=(X.base)]{\node[fill=bestgreen,rounded corners=3pt,inner xsep=3pt,inner ysep=1.5pt](X){\bfseries 0.87};} & \$0.90 & 57.50 & 0.83 & \$1.90 & 100.80 & 0.67 & \$1.55 \\
Q3 & \tikz[baseline=(X.base)]{\node[fill=bestgreen,rounded corners=3pt,inner xsep=3pt,inner ysep=1.5pt](X){\bfseries 8.53};} & 0.78 & \tikz[baseline=(X.base)]{\node[fill=bestgreen,rounded corners=3pt,inner xsep=3pt,inner ysep=1.5pt](X){\bfseries \$2e-3};} & 21.20 & \tikz[baseline=(X.base)]{\node[fill=tiecolor,rounded corners=3pt,inner xsep=3pt,inner ysep=1.5pt](X){\bfseries 0.97};} & \$0.60 & 17.10 & \tikz[baseline=(X.base)]{\node[fill=tiecolor,rounded corners=3pt,inner xsep=3pt,inner ysep=1.5pt](X){\bfseries 0.97};} & \$0.35 & 16.50 & \tikz[baseline=(X.base)]{\node[fill=tiecolor,rounded corners=3pt,inner xsep=3pt,inner ysep=1.5pt](X){\bfseries 0.98};} & \$0.60 & -- & -- & -- \\
\rowcolor{rowgray}
Q4 & \tikz[baseline=(X.base)]{\node[fill=bestgreen,rounded corners=3pt,inner xsep=3pt,inner ysep=1.5pt](X){\bfseries 13.65};} & 0.40 & \tikz[baseline=(X.base)]{\node[fill=bestgreen,rounded corners=3pt,inner xsep=3pt,inner ysep=1.5pt](X){\bfseries \$3e-3};} & 31.00 & \tikz[baseline=(X.base)]{\node[fill=bestgreen,rounded corners=3pt,inner xsep=3pt,inner ysep=1.5pt](X){\bfseries 0.69};} & \$1.85 & 156.70 & 0.45 & \$0.70 & 335.00 & 0.53 & \$1.20 & -- & -- & -- \\
Q5 & \tikz[baseline=(X.base)]{\node[fill=bestgreen,rounded corners=3pt,inner xsep=3pt,inner ysep=1.5pt](X){\bfseries 4.48};} & 0.96 & \tikz[baseline=(X.base)]{\node[fill=bestgreen,rounded corners=3pt,inner xsep=3pt,inner ysep=1.5pt](X){\bfseries \$1e-3};} & 25.70 & \tikz[baseline=(X.base)]{\node[fill=tiecolor,rounded corners=3pt,inner xsep=3pt,inner ysep=1.5pt](X){\bfseries 0.98};} & \$0.85 & 7.40 & \tikz[baseline=(X.base)]{\node[fill=tiecolor,rounded corners=3pt,inner xsep=3pt,inner ysep=1.5pt](X){\bfseries 0.99};} & \$0.20 & 6.60 & \tikz[baseline=(X.base)]{\node[fill=tiecolor,rounded corners=3pt,inner xsep=3pt,inner ysep=1.5pt](X){\bfseries 0.98};} & \$0.35 & -- & -- & -- \\
\rowcolor{rowgray}
Q6 & \tikz[baseline=(X.base)]{\node[fill=bestgreen,rounded corners=3pt,inner xsep=3pt,inner ysep=1.5pt](X){\bfseries 13.34};} & 0.67 & \tikz[baseline=(X.base)]{\node[fill=bestgreen,rounded corners=3pt,inner xsep=3pt,inner ysep=1.5pt](X){\bfseries \$3e-3};} & 34.90 & \tikz[baseline=(X.base)]{\node[fill=tiecolor,rounded corners=3pt,inner xsep=3pt,inner ysep=1.5pt](X){\bfseries 0.88};} & \$1.75 & 114.80 & \tikz[baseline=(X.base)]{\node[fill=tiecolor,rounded corners=3pt,inner xsep=3pt,inner ysep=1.5pt](X){\bfseries 0.89};} & \$0.60 & 143.70 & \tikz[baseline=(X.base)]{\node[fill=tiecolor,rounded corners=3pt,inner xsep=3pt,inner ysep=1.5pt](X){\bfseries 0.89};} & \$2.15 & -- & -- & -- \\
Q7 & 222.74 & 0.77 & \tikz[baseline=(X.base)]{\node[fill=bestgreen,rounded corners=3pt,inner xsep=3pt,inner ysep=1.5pt](X){\bfseries \$0.06};} & \tikz[baseline=(X.base)]{\node[fill=bestgreen,rounded corners=3pt,inner xsep=3pt,inner ysep=1.5pt](X){\bfseries 45.40};} & 0.83 & \$4.30 & 199.40 & 0.75 & \$6.65 & 287.60 & \tikz[baseline=(X.base)]{\node[fill=bestgreen,rounded corners=3pt,inner xsep=3pt,inner ysep=1.5pt](X){\bfseries 0.92};} & \$8.95 & 97.70 & 0.51 & \$0.40 \\
\rowcolor{rowgray}
Q8 & 51.63 & 0.50 & \tikz[baseline=(X.base)]{\node[fill=bestgreen,rounded corners=3pt,inner xsep=3pt,inner ysep=1.5pt](X){\bfseries \$0.01};} & 126.20 & 0.29 & \$91.15 & 4.10 & \tikz[baseline=(X.base)]{\node[fill=tiecolor,rounded corners=3pt,inner xsep=3pt,inner ysep=1.5pt](X){\bfseries 1.00};} & \$0.02 & \tikz[baseline=(X.base)]{\node[fill=bestgreen,rounded corners=3pt,inner xsep=3pt,inner ysep=1.5pt](X){\bfseries 3.90};} & \tikz[baseline=(X.base)]{\node[fill=tiecolor,rounded corners=3pt,inner xsep=3pt,inner ysep=1.5pt](X){\bfseries 1.00};} & \$0.05 & 713.00 & 0.00 & \$1.50 \\
Q9 & \tikz[baseline=(X.base)]{\node[fill=bestgreen,rounded corners=3pt,inner xsep=3pt,inner ysep=1.5pt](X){\bfseries 24.43};} & 0.43 & \tikz[baseline=(X.base)]{\node[fill=bestgreen,rounded corners=3pt,inner xsep=3pt,inner ysep=1.5pt](X){\bfseries \$6e-3};} & 48.60 & \tikz[baseline=(X.base)]{\node[fill=bestgreen,rounded corners=3pt,inner xsep=3pt,inner ysep=1.5pt](X){\bfseries 0.58};} & \$1.05 & 243.40 & 0.55 & \$0.30 & 44.90 & 0.49 & \$0.50 & 872.20 & 0.00 & \$1.55 \\
\rowcolor{rowgray}
Q10 & \tikz[baseline=(X.base)]{\node[fill=bestgreen,rounded corners=3pt,inner xsep=3pt,inner ysep=1.5pt](X){\bfseries 10.95};} & 0.00 & \tikz[baseline=(X.base)]{\node[fill=bestgreen,rounded corners=3pt,inner xsep=3pt,inner ysep=1.5pt](X){\bfseries \$3e-3};} & -- & -- & -- & 519.50 & 0.00 & \$0.30 & 1192.50 & \tikz[baseline=(X.base)]{\node[fill=bestgreen,rounded corners=3pt,inner xsep=3pt,inner ysep=1.5pt](X){\bfseries 0.06};} & \$5.10 & -- & -- & -- \\
Q11 & \tikz[baseline=(X.base)]{\node[fill=bestgreen,rounded corners=3pt,inner xsep=3pt,inner ysep=1.5pt](X){\bfseries 57.46};} & 0.25 & \tikz[baseline=(X.base)]{\node[fill=bestgreen,rounded corners=3pt,inner xsep=3pt,inner ysep=1.5pt](X){\bfseries \$0.01};} & -- & -- & -- & 158.70 & \tikz[baseline=(X.base)]{\node[fill=bestgreen,rounded corners=3pt,inner xsep=3pt,inner ysep=1.5pt](X){\bfseries 0.78};} & \$1.35 & 132.40 & 0.73 & \$3.05 & -- & -- & -- \\
\rowcolor{rowgray}
Q12 & \tikz[baseline=(X.base)]{\node[fill=bestgreen,rounded corners=3pt,inner xsep=3pt,inner ysep=1.5pt](X){\bfseries 13.58};} & 0.91 & \tikz[baseline=(X.base)]{\node[fill=bestgreen,rounded corners=3pt,inner xsep=3pt,inner ysep=1.5pt](X){\bfseries \$3e-3};} & 31.10 & \tikz[baseline=(X.base)]{\node[fill=bestgreen,rounded corners=3pt,inner xsep=3pt,inner ysep=1.5pt](X){\bfseries 0.97};} & \$0.50 & 36.40 & 0.60 & \$0.50 & 31.90 & 0.00 & \$0.70 & -- & -- & -- \\
Q13 & 32.99 & 0.51 & \tikz[baseline=(X.base)]{\node[fill=bestgreen,rounded corners=3pt,inner xsep=3pt,inner ysep=1.5pt](X){\bfseries \$8e-3};} & \tikz[baseline=(X.base)]{\node[fill=bestgreen,rounded corners=3pt,inner xsep=3pt,inner ysep=1.5pt](X){\bfseries 22.40};} & 0.70 & \$1.90 & 274.80 & \tikz[baseline=(X.base)]{\node[fill=tiecolor,rounded corners=3pt,inner xsep=3pt,inner ysep=1.5pt](X){\bfseries 0.74};} & \$1.20 & 238.30 & \tikz[baseline=(X.base)]{\node[fill=tiecolor,rounded corners=3pt,inner xsep=3pt,inner ysep=1.5pt](X){\bfseries 0.74};} & \$2.20 & -- & -- & -- \\
\rowcolor{rowgray}
Q14 & \tikz[baseline=(X.base)]{\node[fill=bestgreen,rounded corners=3pt,inner xsep=3pt,inner ysep=1.5pt](X){\bfseries 48.73};} & \tikz[baseline=(X.base)]{\node[fill=tiecolor,rounded corners=3pt,inner xsep=3pt,inner ysep=1.5pt](X){\bfseries 0.88};} & \tikz[baseline=(X.base)]{\node[fill=bestgreen,rounded corners=3pt,inner xsep=3pt,inner ysep=1.5pt](X){\bfseries \$0.01};} & 73.60 & 0.37 & \$21.30 & 178.20 & \tikz[baseline=(X.base)]{\node[fill=tiecolor,rounded corners=3pt,inner xsep=3pt,inner ysep=1.5pt](X){\bfseries 0.87};} & \$1.15 & -- & -- & -- & -- & -- & -- \\
\midrule
\textbf{Avg.} & \tikz[baseline=(X.base)]{\node[fill=bestgreen,rounded corners=3pt,inner xsep=3pt,inner ysep=1.5pt](X){\bfseries 38.13};} & 0.60 & \tikz[baseline=(X.base)]{\node[fill=bestgreen,rounded corners=3pt,inner xsep=3pt,inner ysep=1.5pt](X){\bfseries \$1e-2};} & 44.75 & 0.67 & \$12.10 & 149.20 & \tikz[baseline=(X.base)]{\node[fill=bestgreen,rounded corners=3pt,inner xsep=3pt,inner ysep=1.5pt](X){\bfseries 0.75};} & \$1.04 & 192.54 & 0.70 & \$2.09 & 363.70 & 0.44 & \$1.03 \\
\textbf{Avg.\ Std.} & 0.15 & 0.02 & \$4e-5 & -- & -- & -- & -- & -- & -- & -- & -- & -- & -- & -- & -- \\
\textbf{Wins} & 11 & 0 & 14 & 2 & 3 & 0 & 0 & 2 & 0 & 1 & 2 & 0 & 0 & 0 & 0 \\
\textbf{Total Cost} & & & \tikz[baseline=(X.base)]{\node[fill=bestgreen,rounded corners=3pt,inner xsep=3pt,inner ysep=1.5pt](X){\bfseries \$0.13};} & & & \$145.25 & & & \$14.52 & & & \$27.15 & & & \$5.15 \\
\bottomrule
\end{tabular}

%% file: tables/full_movie.tex
\footnotesize
\label{tab:blendsql_vs_baseline_gemma_e4b}
\begin{tabular}{l rrr | rrr | rrr | rrr | rrr}
\toprule
  & \multicolumn{3}{c|}{\textbf{BlendSQL}} & \multicolumn{3}{c|}{\textbf{BigQuery}} & \multicolumn{3}{c|}{\textbf{LOTUS}} & \multicolumn{3}{c|}{\textbf{Palimpzest}} & \multicolumn{3}{c}{\textbf{ThalamusDB}} \\
\cmidrule(lr){2-4} \cmidrule(lr){5-7} \cmidrule(lr){8-10} \cmidrule(lr){11-13} \cmidrule(l){14-16}
\textbf{Query} & \textbf{Lat.\,(s)} & \textbf{Quality} & \textbf{Cost\,(5\,runs)} & \textbf{Lat.\,(s)} & \textbf{Quality} & \textbf{Cost\,(5\,runs)} & \textbf{Lat.\,(s)} & \textbf{Quality} & \textbf{Cost\,(5\,runs)} & \textbf{Lat.\,(s)} & \textbf{Quality} & \textbf{Cost\,(5\,runs)} & \textbf{Lat.\,(s)} & \textbf{Quality} & \textbf{Cost\,(5\,runs)} \\
\midrule
Q1 & \tikz[baseline=(X.base)]{\node[fill=bestgreen,rounded corners=3pt,inner xsep=3pt,inner ysep=1.5pt](X){\bfseries 0.39};} & \tikz[baseline=(X.base)]{\node[fill=tiecolor,rounded corners=3pt,inner xsep=3pt,inner ysep=1.5pt](X){\bfseries 1.00};} & \tikz[baseline=(X.base)]{\node[fill=tiecolor,rounded corners=3pt,inner xsep=3pt,inner ysep=1.5pt](X){\bfseries \$1e-4};} & 26.30 & \tikz[baseline=(X.base)]{\node[fill=tiecolor,rounded corners=3pt,inner xsep=3pt,inner ysep=1.5pt](X){\bfseries 1.00};} & \$0.25 & 33.10 & \tikz[baseline=(X.base)]{\node[fill=tiecolor,rounded corners=3pt,inner xsep=3pt,inner ysep=1.5pt](X){\bfseries 1.00};} & \$0.45 & 3.80 & \tikz[baseline=(X.base)]{\node[fill=tiecolor,rounded corners=3pt,inner xsep=3pt,inner ysep=1.5pt](X){\bfseries 1.00};} & \tikz[baseline=(X.base)]{\node[fill=tiecolor,rounded corners=3pt,inner xsep=3pt,inner ysep=1.5pt](X){\bfseries \$5e-3};} & 4.20 & 0.95 & \tikz[baseline=(X.base)]{\node[fill=tiecolor,rounded corners=3pt,inner xsep=3pt,inner ysep=1.5pt](X){\bfseries \$2e-3};} \\
\rowcolor{rowgray}
Q2 & \tikz[baseline=(X.base)]{\node[fill=bestgreen,rounded corners=3pt,inner xsep=3pt,inner ysep=1.5pt](X){\bfseries 0.45};} & 0.96 & \tikz[baseline=(X.base)]{\node[fill=bestgreen,rounded corners=3pt,inner xsep=3pt,inner ysep=1.5pt](X){\bfseries \$1e-4};} & 9.50 & \tikz[baseline=(X.base)]{\node[fill=tiecolor,rounded corners=3pt,inner xsep=3pt,inner ysep=1.5pt](X){\bfseries 1.00};} & \$0.01 & 2.10 & \tikz[baseline=(X.base)]{\node[fill=tiecolor,rounded corners=3pt,inner xsep=3pt,inner ysep=1.5pt](X){\bfseries 1.00};} & \$0.05 & 29.70 & \tikz[baseline=(X.base)]{\node[fill=tiecolor,rounded corners=3pt,inner xsep=3pt,inner ysep=1.5pt](X){\bfseries 1.00};} & \$0.05 & 1.90 & 0.92 & \$0.01 \\
Q3 & \tikz[baseline=(X.base)]{\node[fill=bestgreen,rounded corners=3pt,inner xsep=3pt,inner ysep=1.5pt](X){\bfseries 0.48};} & 0.69 & \tikz[baseline=(X.base)]{\node[fill=bestgreen,rounded corners=3pt,inner xsep=3pt,inner ysep=1.5pt](X){\bfseries \$1e-4};} & 11.00 & 0.64 & \$0.01 & 2.10 & 0.64 & \$0.03 & 4.60 & 0.64 & \$0.05 & 3.10 & \tikz[baseline=(X.base)]{\node[fill=bestgreen,rounded corners=3pt,inner xsep=3pt,inner ysep=1.5pt](X){\bfseries 0.74};} & \$0.01 \\
\rowcolor{rowgray}
Q4 & \tikz[baseline=(X.base)]{\node[fill=bestgreen,rounded corners=3pt,inner xsep=3pt,inner ysep=1.5pt](X){\bfseries 0.32};} & 0.64 & \tikz[baseline=(X.base)]{\node[fill=bestgreen,rounded corners=3pt,inner xsep=3pt,inner ysep=1.5pt](X){\bfseries \$8e-5};} & 11.40 & 0.64 & \$0.01 & 2.80 & 0.64 & \$0.03 & 4.40 & \tikz[baseline=(X.base)]{\node[fill=tiecolor,rounded corners=3pt,inner xsep=3pt,inner ysep=1.5pt](X){\bfseries 0.74};} & \$0.10 & 3.80 & \tikz[baseline=(X.base)]{\node[fill=tiecolor,rounded corners=3pt,inner xsep=3pt,inner ysep=1.5pt](X){\bfseries 0.74};} & \$0.01 \\
Q5 & \tikz[baseline=(X.base)]{\node[fill=bestgreen,rounded corners=3pt,inner xsep=3pt,inner ysep=1.5pt](X){\bfseries 0.63};} & 0.46 & \tikz[baseline=(X.base)]{\node[fill=tiecolor,rounded corners=3pt,inner xsep=3pt,inner ysep=1.5pt](X){\bfseries \$2e-4};} & 54.50 & 0.89 & \$5.05 & 536.50 & 0.59 & \$11.90 & 1.90 & 0.39 & \$0.05 & 2.30 & \tikz[baseline=(X.base)]{\node[fill=bestgreen,rounded corners=3pt,inner xsep=3pt,inner ysep=1.5pt](X){\bfseries 1.00};} & \tikz[baseline=(X.base)]{\node[fill=tiecolor,rounded corners=3pt,inner xsep=3pt,inner ysep=1.5pt](X){\bfseries \$5e-3};} \\
\rowcolor{rowgray}
Q6 & \tikz[baseline=(X.base)]{\node[fill=bestgreen,rounded corners=3pt,inner xsep=3pt,inner ysep=1.5pt](X){\bfseries 0.57};} & 0.46 & \tikz[baseline=(X.base)]{\node[fill=tiecolor,rounded corners=3pt,inner xsep=3pt,inner ysep=1.5pt](X){\bfseries \$1e-4};} & 54.50 & 0.69 & \$5.00 & 432.40 & 0.67 & \$9.05 & 2.30 & \tikz[baseline=(X.base)]{\node[fill=tiecolor,rounded corners=3pt,inner xsep=3pt,inner ysep=1.5pt](X){\bfseries 0.83};} & \$0.05 & 1.70 & \tikz[baseline=(X.base)]{\node[fill=tiecolor,rounded corners=3pt,inner xsep=3pt,inner ysep=1.5pt](X){\bfseries 0.84};} & \tikz[baseline=(X.base)]{\node[fill=tiecolor,rounded corners=3pt,inner xsep=3pt,inner ysep=1.5pt](X){\bfseries \$4e-3};} \\
Q7 & \tikz[baseline=(X.base)]{\node[fill=bestgreen,rounded corners=3pt,inner xsep=3pt,inner ysep=1.5pt](X){\bfseries 149.03};} & 0.54 & \tikz[baseline=(X.base)]{\node[fill=bestgreen,rounded corners=3pt,inner xsep=3pt,inner ysep=1.5pt](X){\bfseries \$0.04};} & 198.00 & \tikz[baseline=(X.base)]{\node[fill=tiecolor,rounded corners=3pt,inner xsep=3pt,inner ysep=1.5pt](X){\bfseries 0.70};} & \$16.55 & 431.80 & 0.21 & \$9.05 & 1056.10 & \tikz[baseline=(X.base)]{\node[fill=tiecolor,rounded corners=3pt,inner xsep=3pt,inner ysep=1.5pt](X){\bfseries 0.68};} & \$38.60 & 649.90 & 0.57 & \$0.75 \\
\rowcolor{rowgray}
Q8 & \tikz[baseline=(X.base)]{\node[fill=bestgreen,rounded corners=3pt,inner xsep=3pt,inner ysep=1.5pt](X){\bfseries 0.68};} & \tikz[baseline=(X.base)]{\node[fill=tiecolor,rounded corners=3pt,inner xsep=3pt,inner ysep=1.5pt](X){\bfseries 0.92};} & \tikz[baseline=(X.base)]{\node[fill=bestgreen,rounded corners=3pt,inner xsep=3pt,inner ysep=1.5pt](X){\bfseries \$2e-4};} & 10.90 & 0.76 & \$0.01 & 2.30 & \tikz[baseline=(X.base)]{\node[fill=tiecolor,rounded corners=3pt,inner xsep=3pt,inner ysep=1.5pt](X){\bfseries 0.93};} & \$0.02 & 4.30 & 0.86 & \$0.10 & 6.80 & 0.83 & \$0.03 \\
Q9 & \tikz[baseline=(X.base)]{\node[fill=bestgreen,rounded corners=3pt,inner xsep=3pt,inner ysep=1.5pt](X){\bfseries 0.81};} & \tikz[baseline=(X.base)]{\node[fill=bestgreen,rounded corners=3pt,inner xsep=3pt,inner ysep=1.5pt](X){\bfseries 0.87};} & \tikz[baseline=(X.base)]{\node[fill=bestgreen,rounded corners=3pt,inner xsep=3pt,inner ysep=1.5pt](X){\bfseries \$2e-4};} & 13.30 & 0.78 & \$0.10 & 4.90 & 0.75 & \$0.10 & 5.70 & 0.78 & \$0.25 & -- & -- & -- \\
\rowcolor{rowgray}
Q10 & \tikz[baseline=(X.base)]{\node[fill=bestgreen,rounded corners=3pt,inner xsep=3pt,inner ysep=1.5pt](X){\bfseries 8.88};} & \tikz[baseline=(X.base)]{\node[fill=bestgreen,rounded corners=3pt,inner xsep=3pt,inner ysep=1.5pt](X){\bfseries 0.71};} & \tikz[baseline=(X.base)]{\node[fill=bestgreen,rounded corners=3pt,inner xsep=3pt,inner ysep=1.5pt](X){\bfseries \$2e-3};} & 32.10 & 0.44 & \$0.65 & 30.90 & 0.40 & \$0.65 & 39.20 & 0.42 & \$1.90 & -- & -- & -- \\
\midrule
\textbf{Avg.} & \tikz[baseline=(X.base)]{\node[fill=bestgreen,rounded corners=3pt,inner xsep=3pt,inner ysep=1.5pt](X){\bfseries 16.22};} & 0.72 & \tikz[baseline=(X.base)]{\node[fill=bestgreen,rounded corners=3pt,inner xsep=3pt,inner ysep=1.5pt](X){\bfseries \$4e-3};} & 42.15 & 0.75 & \$2.77 & 147.89 & 0.68 & \$3.13 & 115.20 & 0.73 & \$4.12 & 84.21 & \tikz[baseline=(X.base)]{\node[fill=bestgreen,rounded corners=3pt,inner xsep=3pt,inner ysep=1.5pt](X){\bfseries 0.82};} & \$0.10 \\
\textbf{Avg.\ Std.} & 0.22 & 0.03 & \$6e-5 & -- & -- & -- & -- & -- & -- & -- & -- & -- & -- & -- & -- \\
\textbf{Wins} & 10 & 2 & 7 & 0 & 0 & 0 & 0 & 0 & 0 & 0 & 0 & 0 & 0 & 2 & 0 \\
\textbf{Total Cost} & & & \tikz[baseline=(X.base)]{\node[fill=bestgreen,rounded corners=3pt,inner xsep=3pt,inner ysep=1.5pt](X){\bfseries \$0.04};} & & & \$27.66 & & & \$31.32 & & & \$41.16 & & & \$0.82 \\
\bottomrule
\end{tabular}

%% file: tables/full_cars.tex
\footnotesize
\label{tab:blendsql_vs_baseline_gemma_e4b}
\begin{tabular}{l rrr | rrr | rrr | rrr | rrr}
\toprule
  & \multicolumn{3}{c|}{\textbf{BlendSQL}} & \multicolumn{3}{c|}{\textbf{BigQuery}} & \multicolumn{3}{c|}{\textbf{LOTUS}} & \multicolumn{3}{c|}{\textbf{Palimpzest}} & \multicolumn{3}{c}{\textbf{ThalamusDB}} \\
\cmidrule(lr){2-4} \cmidrule(lr){5-7} \cmidrule(lr){8-10} \cmidrule(lr){11-13} \cmidrule(l){14-16}
\textbf{Query} & \textbf{Lat.\,(s)} & \textbf{Quality} & \textbf{Cost\,(5\,runs)} & \textbf{Lat.\,(s)} & \textbf{Quality} & \textbf{Cost\,(5\,runs)} & \textbf{Lat.\,(s)} & \textbf{Quality} & \textbf{Cost\,(5\,runs)} & \textbf{Lat.\,(s)} & \textbf{Quality} & \textbf{Cost\,(5\,runs)} & \textbf{Lat.\,(s)} & \textbf{Quality} & \textbf{Cost\,(5\,runs)} \\
\midrule
Q1 & 201.92 & 0.63 & \tikz[baseline=(X.base)]{\node[fill=bestgreen,rounded corners=3pt,inner xsep=3pt,inner ysep=1.5pt](X){\bfseries \$0.05};} & \tikz[baseline=(X.base)]{\node[fill=bestgreen,rounded corners=3pt,inner xsep=3pt,inner ysep=1.5pt](X){\bfseries 61.70};} & 0.71 & \$7.20 & 550.00 & \tikz[baseline=(X.base)]{\node[fill=bestgreen,rounded corners=3pt,inner xsep=3pt,inner ysep=1.5pt](X){\bfseries 0.90};} & \$8.70 & 465.60 & 0.69 & \$12.20 & 829.70 & 0.81 & \$6.85 \\
\rowcolor{rowgray}
Q2 & \tikz[baseline=(X.base)]{\node[fill=bestgreen,rounded corners=3pt,inner xsep=3pt,inner ysep=1.5pt](X){\bfseries 1.97};} & 0.00 & \tikz[baseline=(X.base)]{\node[fill=bestgreen,rounded corners=3pt,inner xsep=3pt,inner ysep=1.5pt](X){\bfseries \$5e-4};} & 14.10 & \tikz[baseline=(X.base)]{\node[fill=tiecolor,rounded corners=3pt,inner xsep=3pt,inner ysep=1.5pt](X){\bfseries 0.08};} & \$0.05 & -- & -- & -- & 4.10 & 0.00 & \$0.02 & 14.10 & \tikz[baseline=(X.base)]{\node[fill=tiecolor,rounded corners=3pt,inner xsep=3pt,inner ysep=1.5pt](X){\bfseries 0.09};} & \$0.05 \\
Q3 & \tikz[baseline=(X.base)]{\node[fill=bestgreen,rounded corners=3pt,inner xsep=3pt,inner ysep=1.5pt](X){\bfseries 1.69};} & 0.94 & \tikz[baseline=(X.base)]{\node[fill=bestgreen,rounded corners=3pt,inner xsep=3pt,inner ysep=1.5pt](X){\bfseries \$4e-4};} & 36.40 & \tikz[baseline=(X.base)]{\node[fill=bestgreen,rounded corners=3pt,inner xsep=3pt,inner ysep=1.5pt](X){\bfseries 1.00};} & \$8.30 & 456.20 & 0.90 & \$3.00 & 6.10 & 0.92 & \$0.05 & -- & -- & -- \\
\rowcolor{rowgray}
Q4 & 193.15 & \tikz[baseline=(X.base)]{\node[fill=tiecolor,rounded corners=3pt,inner xsep=3pt,inner ysep=1.5pt](X){\bfseries 0.99};} & \tikz[baseline=(X.base)]{\node[fill=bestgreen,rounded corners=3pt,inner xsep=3pt,inner ysep=1.5pt](X){\bfseries \$0.05};} & \tikz[baseline=(X.base)]{\node[fill=bestgreen,rounded corners=3pt,inner xsep=3pt,inner ysep=1.5pt](X){\bfseries 68.70};} & \tikz[baseline=(X.base)]{\node[fill=tiecolor,rounded corners=3pt,inner xsep=3pt,inner ysep=1.5pt](X){\bfseries 0.99};} & \$7.05 & 822.00 & \tikz[baseline=(X.base)]{\node[fill=tiecolor,rounded corners=3pt,inner xsep=3pt,inner ysep=1.5pt](X){\bfseries 0.99};} & \$8.55 & 443.60 & \tikz[baseline=(X.base)]{\node[fill=tiecolor,rounded corners=3pt,inner xsep=3pt,inner ysep=1.5pt](X){\bfseries 0.99};} & \$12.05 & 768.80 & \tikz[baseline=(X.base)]{\node[fill=tiecolor,rounded corners=3pt,inner xsep=3pt,inner ysep=1.5pt](X){\bfseries 1.00};} & \$6.70 \\
Q5 & \tikz[baseline=(X.base)]{\node[fill=bestgreen,rounded corners=3pt,inner xsep=3pt,inner ysep=1.5pt](X){\bfseries 2.25};} & 0.00 & \tikz[baseline=(X.base)]{\node[fill=bestgreen,rounded corners=3pt,inner xsep=3pt,inner ysep=1.5pt](X){\bfseries \$6e-4};} & 58.90 & \tikz[baseline=(X.base)]{\node[fill=tiecolor,rounded corners=3pt,inner xsep=3pt,inner ysep=1.5pt](X){\bfseries 1.00};} & \$7.35 & -- & -- & -- & 6.30 & \tikz[baseline=(X.base)]{\node[fill=tiecolor,rounded corners=3pt,inner xsep=3pt,inner ysep=1.5pt](X){\bfseries 1.00};} & \$0.05 & 483.70 & \tikz[baseline=(X.base)]{\node[fill=tiecolor,rounded corners=3pt,inner xsep=3pt,inner ysep=1.5pt](X){\bfseries 1.00};} & \$8.05 \\
\rowcolor{rowgray}
Q6 & 219.79 & 0.75 & \tikz[baseline=(X.base)]{\node[fill=bestgreen,rounded corners=3pt,inner xsep=3pt,inner ysep=1.5pt](X){\bfseries \$0.05};} & \tikz[baseline=(X.base)]{\node[fill=bestgreen,rounded corners=3pt,inner xsep=3pt,inner ysep=1.5pt](X){\bfseries 44.30};} & \tikz[baseline=(X.base)]{\node[fill=tiecolor,rounded corners=3pt,inner xsep=3pt,inner ysep=1.5pt](X){\bfseries 0.96};} & \$10.00 & -- & -- & -- & 427.30 & \tikz[baseline=(X.base)]{\node[fill=tiecolor,rounded corners=3pt,inner xsep=3pt,inner ysep=1.5pt](X){\bfseries 0.96};} & \$12.55 & 775.40 & \tikz[baseline=(X.base)]{\node[fill=tiecolor,rounded corners=3pt,inner xsep=3pt,inner ysep=1.5pt](X){\bfseries 0.97};} & \$9.80 \\
Q7 & 378.20 & 0.55 & \tikz[baseline=(X.base)]{\node[fill=bestgreen,rounded corners=3pt,inner xsep=3pt,inner ysep=1.5pt](X){\bfseries \$0.09};} & \tikz[baseline=(X.base)]{\node[fill=bestgreen,rounded corners=3pt,inner xsep=3pt,inner ysep=1.5pt](X){\bfseries 86.00};} & 0.45 & \$15.85 & -- & -- & -- & 882.70 & \tikz[baseline=(X.base)]{\node[fill=tiecolor,rounded corners=3pt,inner xsep=3pt,inner ysep=1.5pt](X){\bfseries 0.56};} & \$22.35 & 2146.20 & \tikz[baseline=(X.base)]{\node[fill=tiecolor,rounded corners=3pt,inner xsep=3pt,inner ysep=1.5pt](X){\bfseries 0.58};} & \$15.30 \\
\rowcolor{rowgray}
Q8 & 173.18 & 0.07 & \tikz[baseline=(X.base)]{\node[fill=bestgreen,rounded corners=3pt,inner xsep=3pt,inner ysep=1.5pt](X){\bfseries \$0.04};} & \tikz[baseline=(X.base)]{\node[fill=bestgreen,rounded corners=3pt,inner xsep=3pt,inner ysep=1.5pt](X){\bfseries 38.20};} & 0.24 & \$8.45 & 1349.80 & \tikz[baseline=(X.base)]{\node[fill=bestgreen,rounded corners=3pt,inner xsep=3pt,inner ysep=1.5pt](X){\bfseries 0.45};} & \$8.90 & 268.70 & 0.29 & \$10.10 & 68.00 & 0.20 & \$1.05 \\
Q10 & 350.62 & 0.50 & \tikz[baseline=(X.base)]{\node[fill=bestgreen,rounded corners=3pt,inner xsep=3pt,inner ysep=1.5pt](X){\bfseries \$0.09};} & \tikz[baseline=(X.base)]{\node[fill=bestgreen,rounded corners=3pt,inner xsep=3pt,inner ysep=1.5pt](X){\bfseries 62.00};} & \tikz[baseline=(X.base)]{\node[fill=bestgreen,rounded corners=3pt,inner xsep=3pt,inner ysep=1.5pt](X){\bfseries 0.57};} & \$13.50 & 618.10 & 0.41 & \$15.45 & 594.90 & 0.51 & \$23.45 & -- & -- & -- \\
\midrule
\textbf{Avg.} & 169.20 & 0.49 & \tikz[baseline=(X.base)]{\node[fill=bestgreen,rounded corners=3pt,inner xsep=3pt,inner ysep=1.5pt](X){\bfseries \$0.04};} & \tikz[baseline=(X.base)]{\node[fill=bestgreen,rounded corners=3pt,inner xsep=3pt,inner ysep=1.5pt](X){\bfseries 52.26};} & 0.67 & \$8.64 & 759.22 & \tikz[baseline=(X.base)]{\node[fill=bestgreen,rounded corners=3pt,inner xsep=3pt,inner ysep=1.5pt](X){\bfseries 0.73};} & \$8.92 & 344.37 & 0.66 & \$10.31 & 726.56 & 0.66 & \$6.83 \\
\textbf{Avg.\ Std.} & 0.10 & 0.01 & \$3e-5 & -- & -- & -- & -- & -- & -- & -- & -- & -- & -- & -- & -- \\
\textbf{Wins} & 3 & 0 & 9 & 6 & 2 & 0 & 0 & 2 & 0 & 0 & 0 & 0 & 0 & 0 & 0 \\
\textbf{Total Cost} & & & \tikz[baseline=(X.base)]{\node[fill=bestgreen,rounded corners=3pt,inner xsep=3pt,inner ysep=1.5pt](X){\bfseries \$0.38};} & & & \$77.75 & & & \$44.60 & & & \$92.82 & & & \$47.80 \\
\bottomrule
\end{tabular}

%% file: tables/full_wildlife.tex
\footnotesize
\label{tab:blendsql_vs_baseline_gemma_e4b}
\begin{tabular}{l rrr | rrr | rrr | rrr | rrr}
\toprule
  & \multicolumn{3}{c|}{\textbf{BlendSQL}} & \multicolumn{3}{c|}{\textbf{BigQuery}} & \multicolumn{3}{c|}{\textbf{LOTUS}} & \multicolumn{3}{c|}{\textbf{Palimpzest}} & \multicolumn{3}{c}{\textbf{ThalamusDB}} \\
\cmidrule(lr){2-4} \cmidrule(lr){5-7} \cmidrule(lr){8-10} \cmidrule(lr){11-13} \cmidrule(l){14-16}
\textbf{Query} & \textbf{Lat.\,(s)} & \textbf{Quality} & \textbf{Cost\,(5\,runs)} & \textbf{Lat.\,(s)} & \textbf{Quality} & \textbf{Cost\,(5\,runs)} & \textbf{Lat.\,(s)} & \textbf{Quality} & \textbf{Cost\,(5\,runs)} & \textbf{Lat.\,(s)} & \textbf{Quality} & \textbf{Cost\,(5\,runs)} & \textbf{Lat.\,(s)} & \textbf{Quality} & \textbf{Cost\,(5\,runs)} \\
\midrule
Q1 & \tikz[baseline=(X.base)]{\node[fill=bestgreen,rounded corners=3pt,inner xsep=3pt,inner ysep=1.5pt](X){\bfseries 9.41};} & \tikz[baseline=(X.base)]{\node[fill=tiecolor,rounded corners=3pt,inner xsep=3pt,inner ysep=1.5pt](X){\bfseries 0.78};} & \tikz[baseline=(X.base)]{\node[fill=bestgreen,rounded corners=3pt,inner xsep=3pt,inner ysep=1.5pt](X){\bfseries \$2e-3};} & 32.00 & \tikz[baseline=(X.base)]{\node[fill=tiecolor,rounded corners=3pt,inner xsep=3pt,inner ysep=1.5pt](X){\bfseries 0.79};} & \$0.55 & 92.40 & \tikz[baseline=(X.base)]{\node[fill=tiecolor,rounded corners=3pt,inner xsep=3pt,inner ysep=1.5pt](X){\bfseries 0.79};} & \$0.55 & 32.80 & \tikz[baseline=(X.base)]{\node[fill=tiecolor,rounded corners=3pt,inner xsep=3pt,inner ysep=1.5pt](X){\bfseries 0.79};} & \$0.65 & 19.60 & \tikz[baseline=(X.base)]{\node[fill=tiecolor,rounded corners=3pt,inner xsep=3pt,inner ysep=1.5pt](X){\bfseries 0.79};} & \$0.55 \\
\rowcolor{rowgray}
Q2 & 4.16 & 0.00 & \tikz[baseline=(X.base)]{\node[fill=bestgreen,rounded corners=3pt,inner xsep=3pt,inner ysep=1.5pt](X){\bfseries \$1e-3};} & 9.40 & \tikz[baseline=(X.base)]{\node[fill=tiecolor,rounded corners=3pt,inner xsep=3pt,inner ysep=1.5pt](X){\bfseries 0.19};} & \$0.05 & -- & -- & -- & \tikz[baseline=(X.base)]{\node[fill=bestgreen,rounded corners=3pt,inner xsep=3pt,inner ysep=1.5pt](X){\bfseries 2.80};} & \tikz[baseline=(X.base)]{\node[fill=tiecolor,rounded corners=3pt,inner xsep=3pt,inner ysep=1.5pt](X){\bfseries 0.17};} & \$0.05 & 4.50 & 0.14 & \$0.05 \\
Q3 & 9.29 & 0.20 & \tikz[baseline=(X.base)]{\node[fill=bestgreen,rounded corners=3pt,inner xsep=3pt,inner ysep=1.5pt](X){\bfseries \$2e-3};} & 25.70 & 0.00 & \$0.55 & 99.40 & \tikz[baseline=(X.base)]{\node[fill=bestgreen,rounded corners=3pt,inner xsep=3pt,inner ysep=1.5pt](X){\bfseries 1.00};} & \$0.55 & 22.70 & 0.00 & \$0.65 & \tikz[baseline=(X.base)]{\node[fill=bestgreen,rounded corners=3pt,inner xsep=3pt,inner ysep=1.5pt](X){\bfseries 4.30};} & 0.00 & \$0.15 \\
\rowcolor{rowgray}
Q4 & 4.01 & 0.40 & \tikz[baseline=(X.base)]{\node[fill=tiecolor,rounded corners=3pt,inner xsep=3pt,inner ysep=1.5pt](X){\bfseries \$1e-3};} & 9.40 & \tikz[baseline=(X.base)]{\node[fill=bestgreen,rounded corners=3pt,inner xsep=3pt,inner ysep=1.5pt](X){\bfseries 1.00};} & \$0.05 & -- & -- & -- & 2.70 & 0.00 & \$0.05 & \tikz[baseline=(X.base)]{\node[fill=bestgreen,rounded corners=3pt,inner xsep=3pt,inner ysep=1.5pt](X){\bfseries 1.30};} & 0.00 & \tikz[baseline=(X.base)]{\node[fill=tiecolor,rounded corners=3pt,inner xsep=3pt,inner ysep=1.5pt](X){\bfseries \$5e-3};} \\
Q5 & 13.39 & \tikz[baseline=(X.base)]{\node[fill=bestgreen,rounded corners=3pt,inner xsep=3pt,inner ysep=1.5pt](X){\bfseries 0.93};} & \tikz[baseline=(X.base)]{\node[fill=bestgreen,rounded corners=3pt,inner xsep=3pt,inner ysep=1.5pt](X){\bfseries \$3e-3};} & 19.20 & 0.75 & \$0.60 & -- & -- & -- & 13.50 & 0.75 & \$0.65 & \tikz[baseline=(X.base)]{\node[fill=bestgreen,rounded corners=3pt,inner xsep=3pt,inner ysep=1.5pt](X){\bfseries 2.30};} & 0.75 & \$0.05 \\
\rowcolor{rowgray}
Q6 & 13.34 & 0.20 & \tikz[baseline=(X.base)]{\node[fill=bestgreen,rounded corners=3pt,inner xsep=3pt,inner ysep=1.5pt](X){\bfseries \$3e-3};} & 24.30 & 0.20 & \$0.60 & -- & -- & -- & 19.30 & 0.00 & \$0.65 & \tikz[baseline=(X.base)]{\node[fill=bestgreen,rounded corners=3pt,inner xsep=3pt,inner ysep=1.5pt](X){\bfseries 13.20};} & \tikz[baseline=(X.base)]{\node[fill=bestgreen,rounded corners=3pt,inner xsep=3pt,inner ysep=1.5pt](X){\bfseries 0.50};} & \$0.30 \\
Q7 & \tikz[baseline=(X.base)]{\node[fill=bestgreen,rounded corners=3pt,inner xsep=3pt,inner ysep=1.5pt](X){\bfseries 18.58};} & \tikz[baseline=(X.base)]{\node[fill=tiecolor,rounded corners=3pt,inner xsep=3pt,inner ysep=1.5pt](X){\bfseries 1.00};} & \tikz[baseline=(X.base)]{\node[fill=bestgreen,rounded corners=3pt,inner xsep=3pt,inner ysep=1.5pt](X){\bfseries \$5e-3};} & 24.60 & \tikz[baseline=(X.base)]{\node[fill=tiecolor,rounded corners=3pt,inner xsep=3pt,inner ysep=1.5pt](X){\bfseries 1.00};} & \$1.10 & 188.30 & \tikz[baseline=(X.base)]{\node[fill=tiecolor,rounded corners=3pt,inner xsep=3pt,inner ysep=1.5pt](X){\bfseries 1.00};} & \$1.15 & 43.90 & \tikz[baseline=(X.base)]{\node[fill=tiecolor,rounded corners=3pt,inner xsep=3pt,inner ysep=1.5pt](X){\bfseries 1.00};} & \$0.65 & 28.80 & \tikz[baseline=(X.base)]{\node[fill=tiecolor,rounded corners=3pt,inner xsep=3pt,inner ysep=1.5pt](X){\bfseries 1.00};} & \$1.00 \\
\rowcolor{rowgray}
Q8 & 26.68 & \tikz[baseline=(X.base)]{\node[fill=bestgreen,rounded corners=3pt,inner xsep=3pt,inner ysep=1.5pt](X){\bfseries 0.83};} & \tikz[baseline=(X.base)]{\node[fill=bestgreen,rounded corners=3pt,inner xsep=3pt,inner ysep=1.5pt](X){\bfseries \$7e-3};} & 35.50 & 0.75 & \$1.15 & -- & -- & -- & 34.90 & 0.75 & \$0.65 & \tikz[baseline=(X.base)]{\node[fill=bestgreen,rounded corners=3pt,inner xsep=3pt,inner ysep=1.5pt](X){\bfseries 23.80};} & 0.75 & \$0.60 \\
Q9 & \tikz[baseline=(X.base)]{\node[fill=bestgreen,rounded corners=3pt,inner xsep=3pt,inner ysep=1.5pt](X){\bfseries 11.92};} & 0.61 & \tikz[baseline=(X.base)]{\node[fill=bestgreen,rounded corners=3pt,inner xsep=3pt,inner ysep=1.5pt](X){\bfseries \$3e-3};} & 37.60 & 0.59 & \$0.60 & -- & -- & -- & 19.10 & 0.57 & \$0.65 & 17.20 & \tikz[baseline=(X.base)]{\node[fill=bestgreen,rounded corners=3pt,inner xsep=3pt,inner ysep=1.5pt](X){\bfseries 0.67};} & \$0.40 \\
\rowcolor{rowgray}
Q10 & 9.27 & 0.40 & \tikz[baseline=(X.base)]{\node[fill=bestgreen,rounded corners=3pt,inner xsep=3pt,inner ysep=1.5pt](X){\bfseries \$2e-3};} & 40.30 & 0.00 & \$0.55 & 87.80 & \tikz[baseline=(X.base)]{\node[fill=bestgreen,rounded corners=3pt,inner xsep=3pt,inner ysep=1.5pt](X){\bfseries 1.00};} & \$0.55 & 19.60 & 0.00 & \$0.65 & \tikz[baseline=(X.base)]{\node[fill=bestgreen,rounded corners=3pt,inner xsep=3pt,inner ysep=1.5pt](X){\bfseries 4.40};} & 0.00 & \$0.15 \\
\midrule
\textbf{Avg.} & 12.01 & 0.54 & \tikz[baseline=(X.base)]{\node[fill=bestgreen,rounded corners=3pt,inner xsep=3pt,inner ysep=1.5pt](X){\bfseries \$3e-3};} & 25.80 & 0.53 & \$0.58 & 116.98 & \tikz[baseline=(X.base)]{\node[fill=bestgreen,rounded corners=3pt,inner xsep=3pt,inner ysep=1.5pt](X){\bfseries 0.95};} & \$0.70 & 21.13 & 0.40 & \$0.53 & \tikz[baseline=(X.base)]{\node[fill=bestgreen,rounded corners=3pt,inner xsep=3pt,inner ysep=1.5pt](X){\bfseries 11.94};} & 0.46 & \$0.33 \\
\textbf{Avg.\ Std.} & 0.08 & 0.22 & \$2e-5 & -- & -- & -- & -- & -- & -- & -- & -- & -- & -- & -- & -- \\
\textbf{Wins} & 3 & 2 & 9 & 0 & 1 & 0 & 0 & 2 & 0 & 1 & 0 & 0 & 6 & 2 & 0 \\
\textbf{Total Cost} & & & \tikz[baseline=(X.base)]{\node[fill=bestgreen,rounded corners=3pt,inner xsep=3pt,inner ysep=1.5pt](X){\bfseries \$0.03};} & & & \$5.80 & & & \$2.80 & & & \$5.30 & & & \$3.25 \\
\bottomrule
\end{tabular}